\documentclass[10pt,twocolumn,letterpaper]{article}

\usepackage{cvpr}
\usepackage{times}
\usepackage{epsfig}
\usepackage{graphicx}
\usepackage{amsmath}
\usepackage{amssymb}

\usepackage{threeparttable}
\usepackage{epstopdf}
\usepackage{amsfonts}
\usepackage{subfigure}
\usepackage{color}
\usepackage{multirow, multicol, makecell}

\usepackage{algorithm} 
\usepackage{algorithmic} 
\usepackage{threeparttable}
\usepackage{epstopdf}
\usepackage{setspace}
\usepackage{subfigure}
\usepackage{epstopdf}

\usepackage{float}
\usepackage{setspace}

\usepackage[breaklinks=true,bookmarks=false]{hyperref}

\cvprfinalcopy 

\def\cvprPaperID{****} 

\begin{document}

\title{MGANet: A Robust Model for Quality Enhancement of Compressed Video}

\author{%
Xiandong Meng$^{1}$, Xuan Deng$^{2}$, Shuyuan Zhu$^{2}$,\\ Shuaicheng Liu$^{2,3}$, Chuan Wang$^{3}$, Chen Chen$^{4}$, Bing Zeng$^{2}$\\[0.5em]
$^{1}$The Hong Kong University of Science and Technology   $^{3}$Megvii Inc \\
$^{2}$University of Electronic Science and Technology of China    $^{4}$Tencent Research\\
}

\maketitle

\begin{abstract}
In video compression, most of the existing deep learning approaches concentrate on the visual quality of a single frame, while ignoring the useful priors as well as the temporal information of adjacent frames. In this paper, we propose a multi-frame guided attention network (MGANet) to enhance the quality of compressed videos. Our network is composed of a temporal encoder that discovers inter-frame relations, a guided encoder-decoder subnet that encodes and enhances the visual patterns of target-frame, and a multi-supervised  reconstruction component that aggregates information to predict details. We design a bidirectional  residual convolutional LSTM unit to implicitly discover frames variations over time with respect to the target frame. Meanwhile, the guided map is proposed to guide our network to concentrate more on the block boundary. Our approach takes advantage of intra-frame prior information and inter-frame information to improve the quality of compressed video. Experimental results show the robustness and superior performance of the proposed method. Code is available at \url{https://github.com/mengab/MGANet}

\end{abstract}
\section{Introduction}
\begin{figure}[tb]
  \centering
  \includegraphics[width=0.47\textwidth]{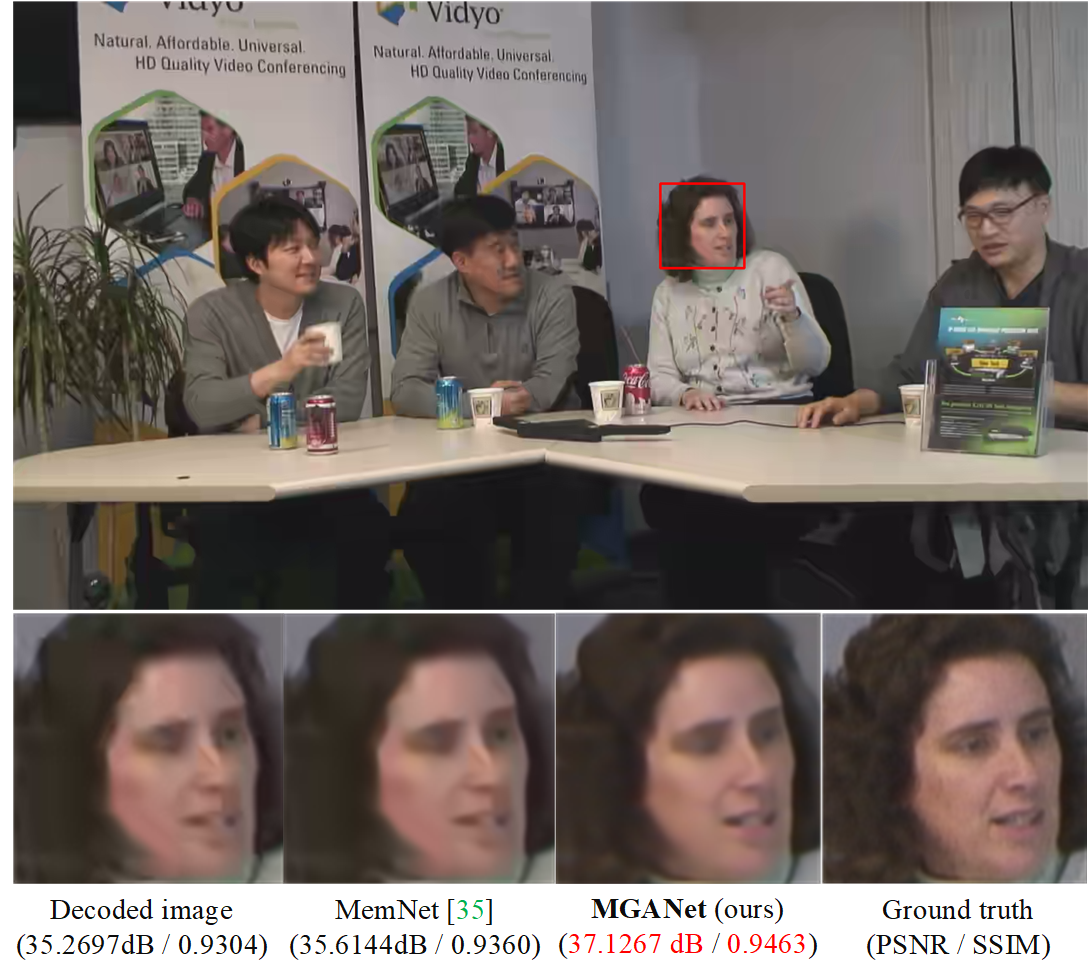}
  \caption{\label{figure1}One real example, the blocking artifacts and quality enhancement of compressed video}
\end{figure}

Uncompressed videos generate a huge quantity of data, for example, without compression, a 90-minutes 8-bits full color high definition movie (1920$\times$1080 pixels per frame) with 30 frames per second occupies 1007.76G Bytes, which is a huge burden for current memory storages or network bandwidth. According to the Cisco Visual Networking Index \cite{Cisco}, more than 75$\%$ of the world’s mobile data traffic will be video by 2021.   As a result, video compression has to be applied to significantly save the coding bit-rate \cite{ Li1,Sullivan1}. However, due to the coarse quantization and motion compensation, many compression artifacts are introduced at low bit-rates \cite{ref5,Zhang18_16}, such as ringing, blurring  and blockiness in boundary regions. As illustrated in Figure \ref{figure1}, the artifacts are characterized by visually noticeable discontinuity. Therefore, video enhancement technique becomes an attractive and promising solution, which can remarkably reduce artifacts to a specific bit rate of compression. The purpose of compression artifacts reduction is to take advantage of the information in compressed bit-stream, to suppress the artifacts and obtain a high-quality reconstruction image. \cite{Zhang18_16}. \\
Traditional image enhancement methods take a single frame with artifacts as input, and usually formulate it as a highly  ill-posed image inverse problem by exploiting some image prior knowledge and observed data  at the decoder \cite{Foi7, Chang8, ref5, kang,  Zhai82}. Most of these methods involve intensive, sometimes heuristic, parameter-tuning and expensive computation. The simplified assumptions on compression noise often hinder their performance on real-word examples. \\
With the success of deep learning in computer vision for image/video super-resolution \cite{Caballero} and video inpainting \cite{Chuan1}, the deep learning based quality enhancement of compressed image/video methods have emerged \cite{p1,Dai14,Dong9_15,Guo10, Jancsary11, Tai17,  Wang15_17Chao, Wang12_16, Yang16}. Among them, Yang \textit {et al.} \cite {Yang16} have achieved state-of-the-art results using a multi-frame quality enhancement network (MFQE). Similar to the video super-resolution methods \cite{Caballero, Unet5, twostep1, twostep2}, MFQE is composed of two steps: a subnet is first used to estimate and compensate the temporal motion across frames, and then  a quality enhancement subnet is used to reduce the compression artifacts.
Despite the success of MFQE approach, we still find its limitations in two aspects: 1)  The results of this  two-step method rely  on the accuracy of motion estimation (ME). If the ME is not accurate, we have found by experiments that the compensation operation actually causes more serious interference to the target frame; 2) Motion is only one kind of temporal information, other information such as brightness or color variations is not well explored in this kind of two-step method.

Different from previous methods, in this paper, we explore a more effective network structure for quality enhancement of compressed video by fully take advantage of the intra-frame prior information and inter-frame information. Instead of explicitly calculating and compensating for motion between input frames, the proposed bidirectional residual convolutional LSTM unit implicitly explores information that is beneficial to the reconstructed output frames. We generate the guided map by the partition information of Transform Unit (TU) in the High Efficiency Video Coding (HEVC) to guide our proposed network to concentrate more on the block boundary. The guided map is fused into our MGANet by a guided attention encoder-decoder module, which is a two-channel encoder subnet with shared network weights. Finally, the reconstruction video is generated by training a multi-supervised loss function.

The main contributions of this paper are: (1) As the first attempt on multi-frame quality enhancement of compressed video utilizing the prior partition information of transform unit, as shown in Figure \ref{figure1} and to be elaborated later, the proposed method can produce better quality results than other state-of-the-art  approaches, which also opens up new space for exploring the quality enhancement of compressed video in the future. (2) Our proposed method greatly improves the robustness of the network by fully take advantage of the intra-frame prior information and inter-frame information. (3) We establish a training database for TUs' partition based on HEVC at both intra- and inter-modes, which may facilitate the applications of this prior information in quality enhancement of compression video.
\section{Related Work}
In general, image/video quality enhancement methods can be divided into two categories: single-frame approches and multi-frame approaches. For the single-frame approaches, Yoo \textit {et al.} first proposed to classify the compressed image into smooth region and  edge region, and developed a two-step framework for reducing blocking artifacts in different regions based on inter-block correlation \cite{Yoo6}.
Then, Foi et al. built a shape-adaptive discrete cosine transform (DCT)  model to reduce the artifacts that caused by compression \cite{Foi}.
Recently, Dong \textit {et al.} \cite{Dong9_15} proposed ARCNN to reduce the JPEG artifacts of images. Later, DnCNN \cite{Zhang13_17} and MemNet \cite{Tai17} were proposed for several tasks of image restoration, including quality enhancement. For the quality enhancement of video compression, VRCNN \cite{Dai14} was proposed as a variable-filter-size residue-learning convolutional neural networks for the post processing of HEVC intra coding. Afterwards, He \textit {et al.} \cite{yao1} utilized the CU's partition information produced by the encoder to guide the quality enhancement process. Wang et al. \cite{Wang15_17Chao} developed a Deep CNN-based Auto Decoder (DCAD), which contains 10 CNN layers to reduce the distortion of compressed video.  Although these single-frame methods are effective in enhancing  performance of single video frame,  the overall performance improvement of compressed video is limited.
Most recently, Yang \textit {et al.} \cite{Yang16} proposed a MFQE model with multi-frame input  for quality enhancement of compressed video by considering the information of neighboring frames. However, we find the results of this method rely on the accuracy of motion estimation, and the accuracy of motion estimation for compressed video is also a challenge task.
\section{\label{section3}Source Analysis of Compression Artifact.}  
\begin{figure*}[!htp]
\centering
\subfigure[$\rm{T}^{\rm{th}}$ decoded frame, size of 832$\times$480]{
\begin{minipage}[t]{0.32\textwidth}
\centering
\includegraphics[width=0.93\textwidth]{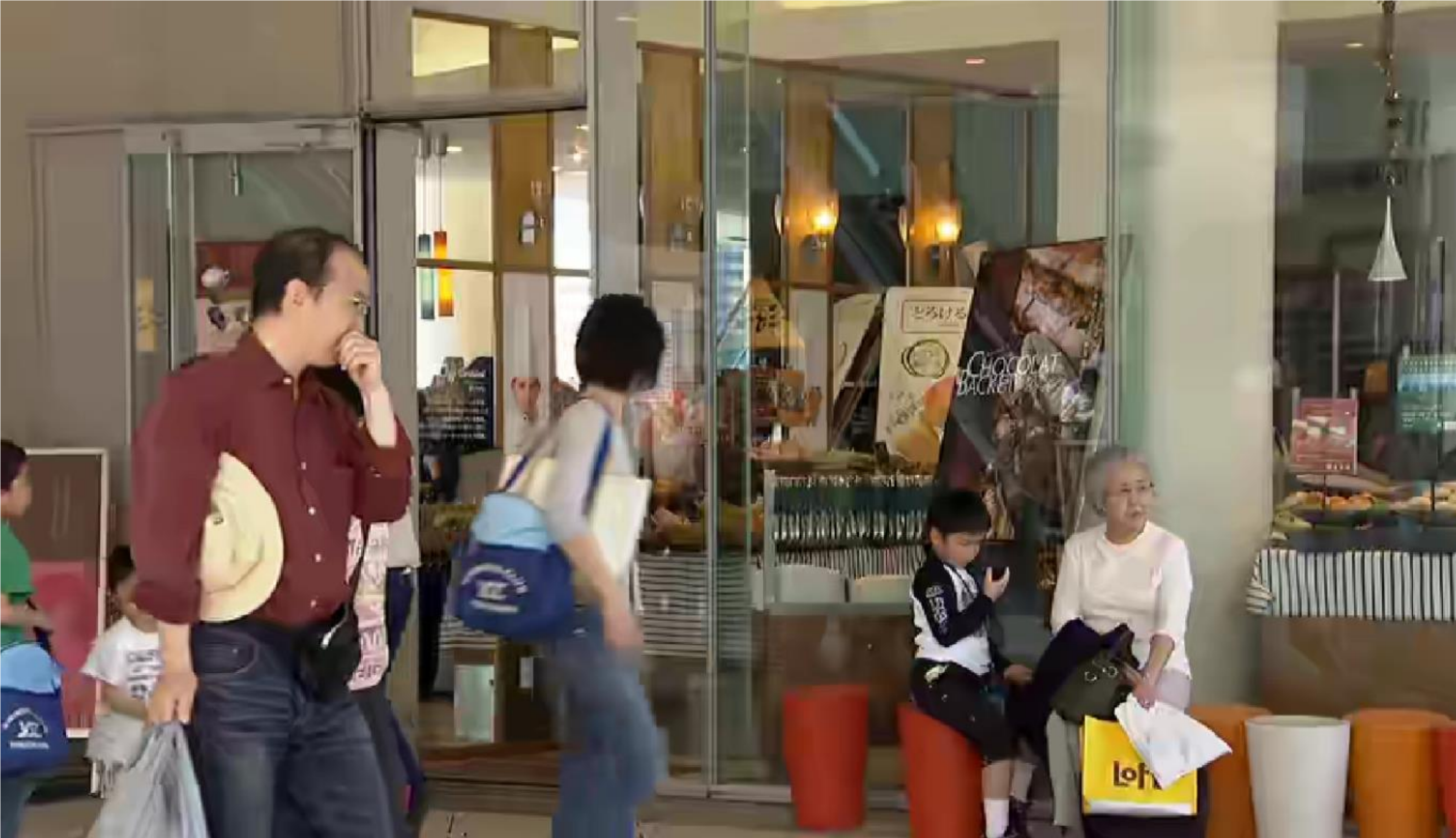}
\end{minipage}
}
\subfigure[standard deviation image of (a)]{
\begin{minipage}[t]{0.317\textwidth}
\centering
\includegraphics[width=1.02\textwidth]{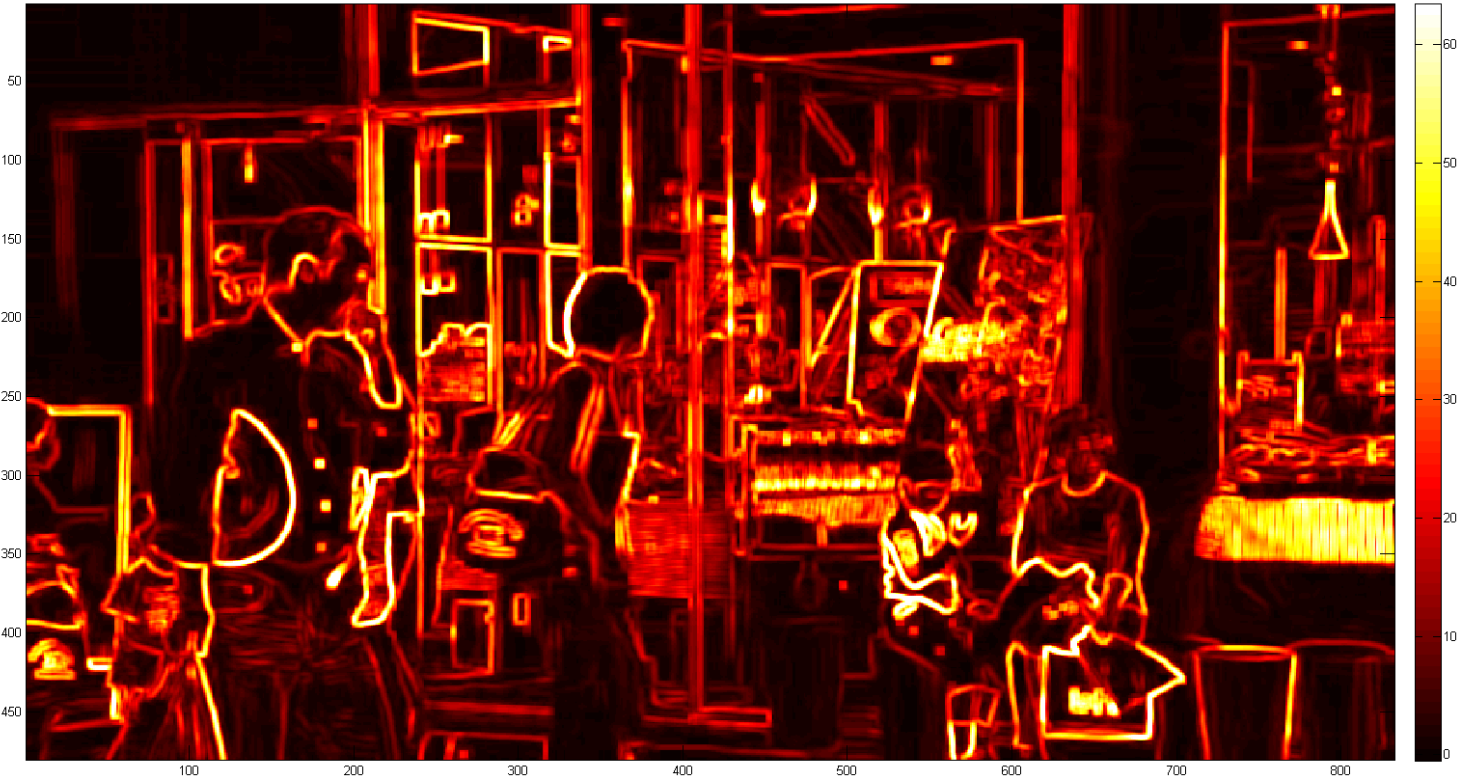}
\end{minipage}
}
\subfigure[Spatial noise difference, $\rm{T}^{\rm{th}}$]{
\begin{minipage}[t]{0.317\textwidth}
\centering
\includegraphics[width=1.02\textwidth]{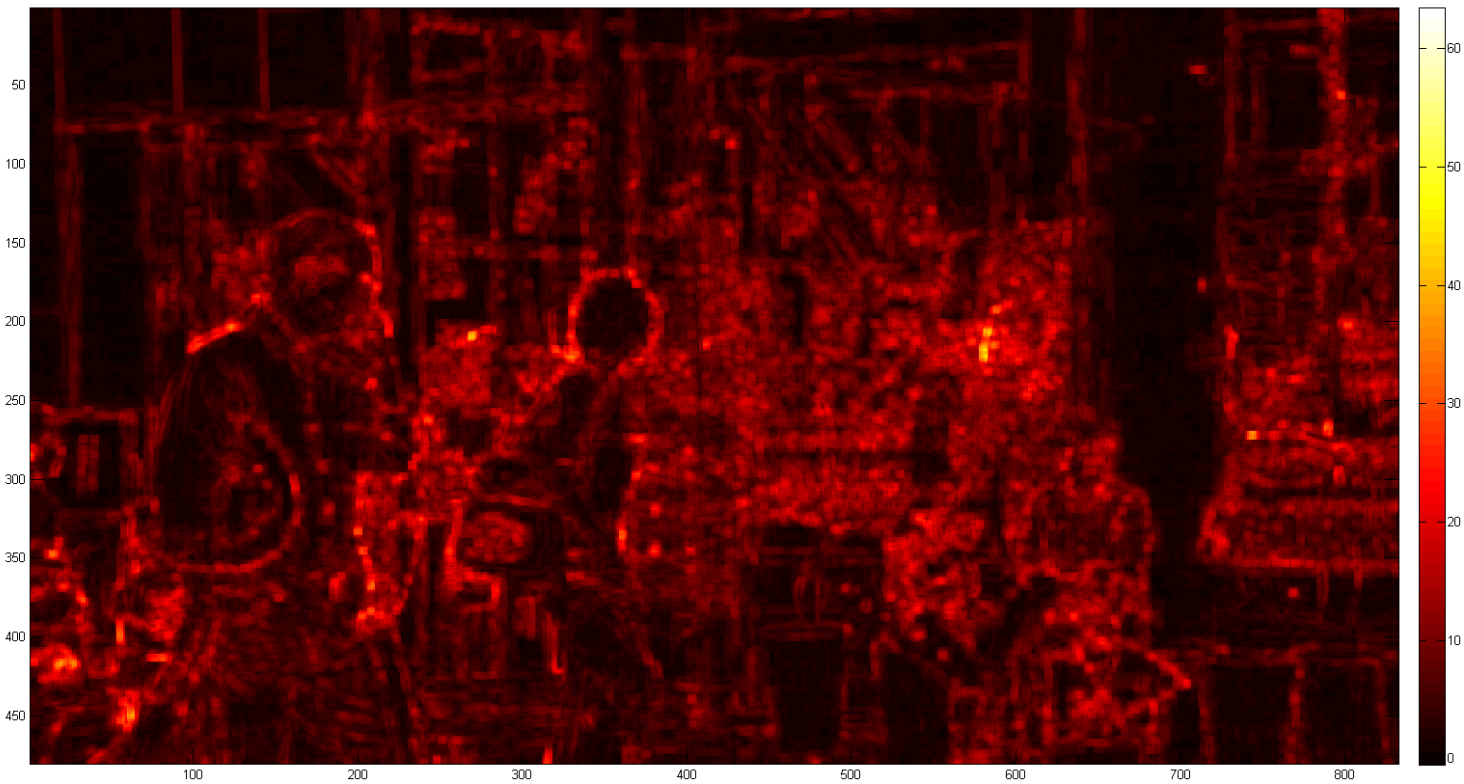}
\end{minipage}
}\\
\subfigure[Temporal noise difference, $\rm{T}^{\rm{th}}$ $\to$$\rm{T}^{\rm{th}}$$\rm{+1}$ ]{
\begin{minipage}[t]{0.32\textwidth}
\centering
\includegraphics[width=1\textwidth]{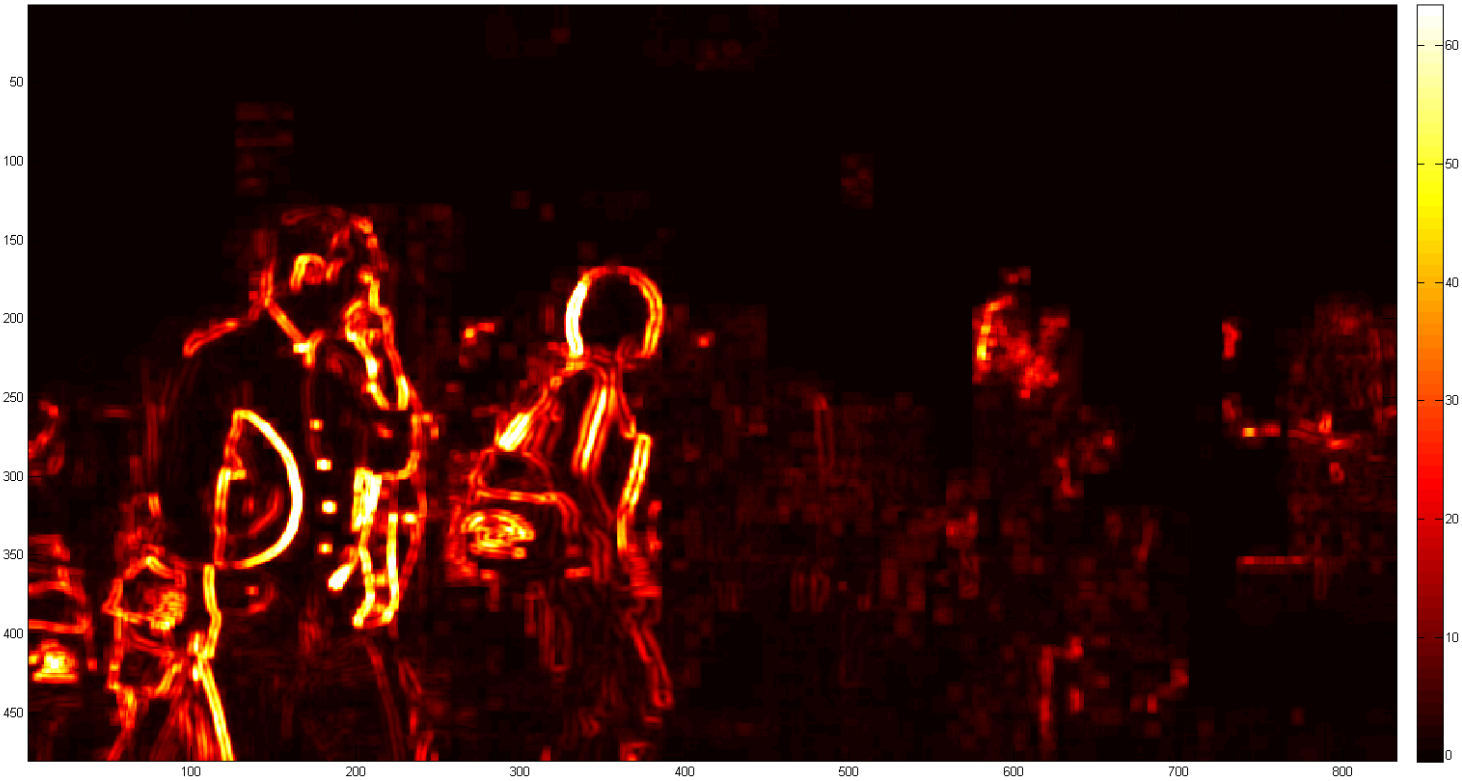}
\end{minipage}
}
\subfigure[Temporal noise difference, $\rm{T}^{\rm{th}}$ $\to$$\rm{T}^{\rm{th}}$$\rm{+2}$ ]{
\begin{minipage}[t]{0.32\textwidth}
\centering
\includegraphics[width=1\textwidth]{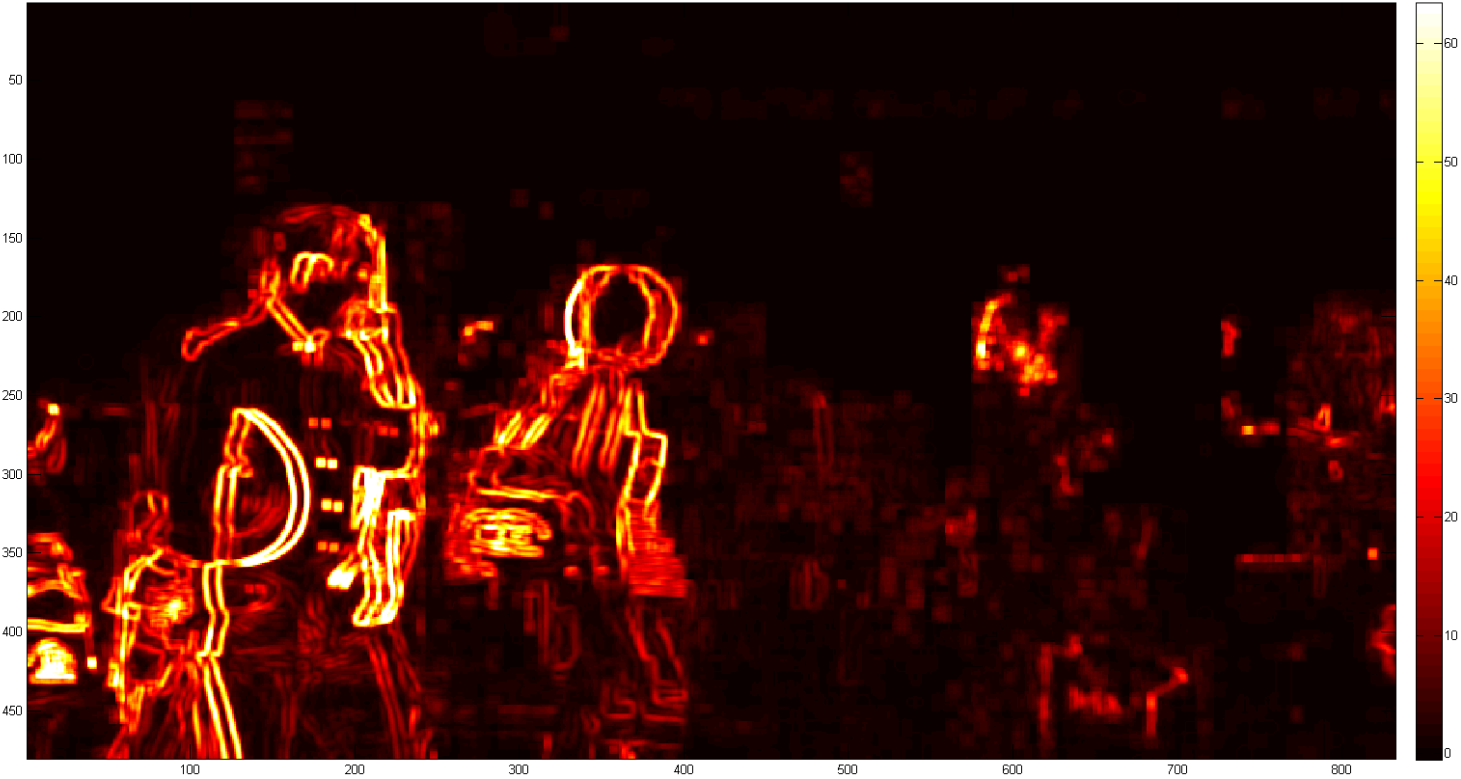}
\end{minipage}
}
\subfigure[Temporal noise difference, $\rm{T}^{\rm{th}}$ $\to$$\rm{T}^{\rm{th}}$$\rm{+3}$ ]{
\begin{minipage}[t]{0.32\textwidth}
\centering
\includegraphics[width=1\textwidth]{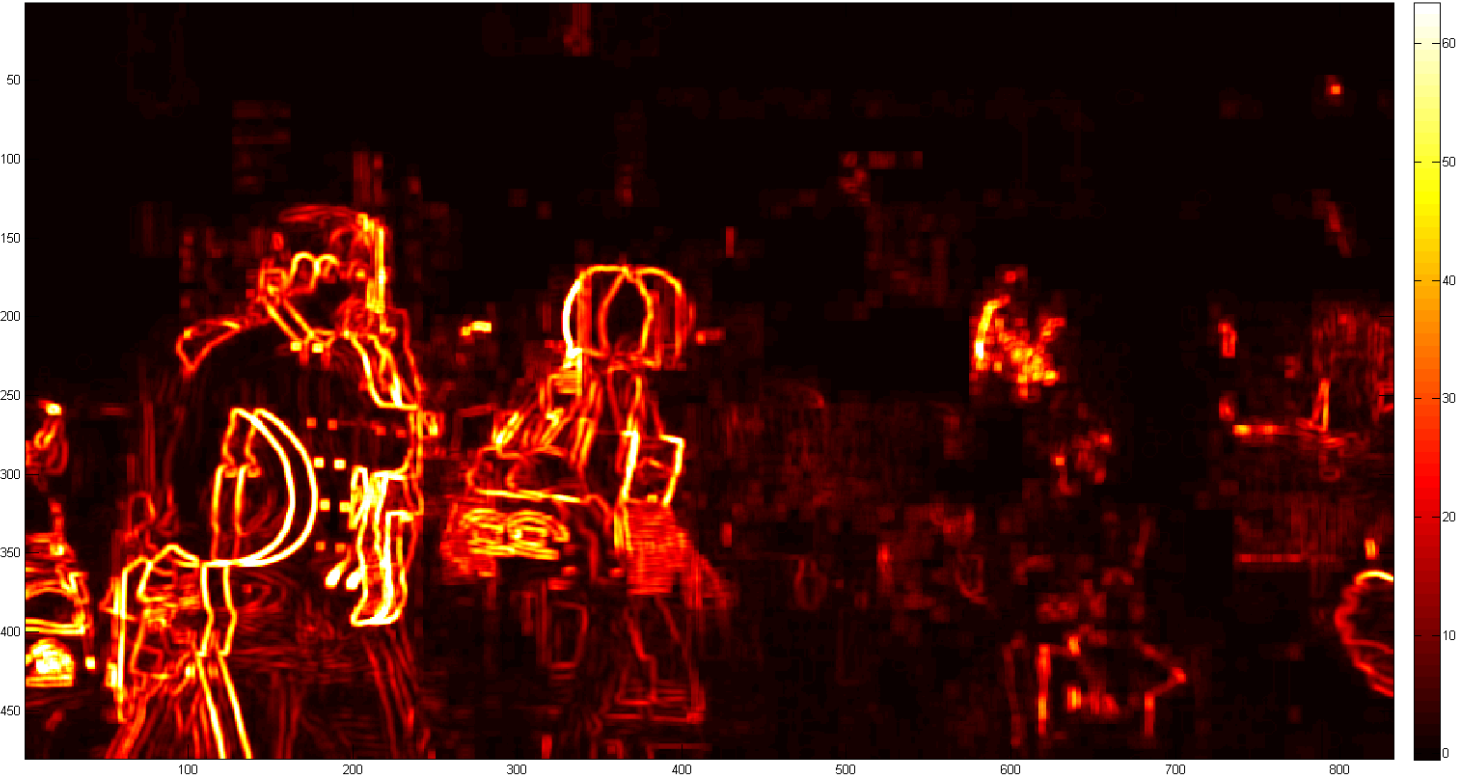}
\end{minipage}
}
\centering
\caption{ \label{figure2}Standard deviation image of  compressed noise difference in spatial-temporal domain}
\end{figure*}
Typically, in most popular lossy compression schemes, such as H.264/AVC \cite{Wiegand2}, HEVC \cite{Sullivan1} and VVC \cite{VVC}, the compression process normally consists of five steps. a) split the input frame into small blocks; b) intra/inter-frame prediction; c) apply DCT on prediction block; d) divide DCT coefficients of each block by quantization parameters, and round the quantized values; e) use entropy coding to generate the compressed bit-streams.  Due to the independent quantization of DCT coefficients in each block, the information loss for the compression only takes place in the step of quantization. Take the HEVC as an example, suppose {\bf{x}} is the original  image of size $N \times N$, let {\bf{y}} be the decompressed image, and denote $ \bf{\hat x}$ and $\bf{\hat y}$ as the frequency images of  {\bf{x}} and  {\bf{y}}, respectively. Then,
\begin{equation}
{\bf{\hat x}} = \cal{rT}{(\bf{x})};    {\kern 10pt} {\bf{\hat y}} =\cal{rT}{(\bf{y})}
\end{equation}
where $ \cal{rT}$ denotes DCT operation. According to the process of  compression described above, we have
\begin{equation}\label{eq2}
{{{\bf{\hat y}}}_{[(k - 1) \times N + l]}} = R\left( {\frac{{{{{\bf{\hat x}}}_{[(k - 1) \times N + l]}}}}{{{\bf{Q}}_{[k,l]}^p}}} \right) \times {\bf{Q}}_{[k,l]}^p
\end{equation}
where ${\rm{1}}\le k,{\kern 2pt}  l\le N$, $R$ is the round function, ${\bf{Q}}^p$ is the quantization matrix of size $p \times p$, $p \in \left\{ {4,8,16,32} \right\}$. From equation (\ref{eq2}), we can see that the rounding operation will result in the loss of high-frequency coefficients of transform block, the larger the ${\bf{Q}}^p$,  the more high frequency coefficients will be lost.  As a result, the blocking artifacts are normally characterized by visually noticeable discontinuity between neighboring blocks, especially at low bit-rate. 

Figure \ref{figure2}{ (b)}  and Figure \ref{figure2}{ (c)} show the standard deviations of decoded frame and the quantization noise for the $T-th$ frame in a compressed video by HEVC intra coding at QP 37. The standard deviation of compression noise for every decoded pixel is calculated in a 5$\times$5 neighborhood centered at the corresponding decoded pixel. The bright areas indicate large quantization noise. Conversely, dark areas indicate small quantization noise. From Figure \ref{figure2}{ (c)}, we can see that the compression noise level varies significantly with different image contents, and higher noise levels usually distribute around image edges and texture regions, while lower noise levels usually exist in smooth areas. Figures \ref{figure2}{ (d)} - \ref{figure2}{ (f)} show the standard deviation of the target frame quantization noise over time from the quantization noise of adjacent frames. The bright area indicates a large difference between the adjacent frames. Due to motion compensation is used for inter-frame, as shown in figures \ref{figure2}{ (d)} - \ref{figure2}{ (f)}, the compression noise is also noticeable discontinuity between neighboring frames for moving objects.  As the temporal neighborhood radius increases, this discontinuity also gradually increases. This means that if we use inter-frame prior information, we need to design a module in the network that extracts useful information from adjacent frames and filters out interference information.

\begin{figure}[!htp]
 \center{\includegraphics[width=70mm]  {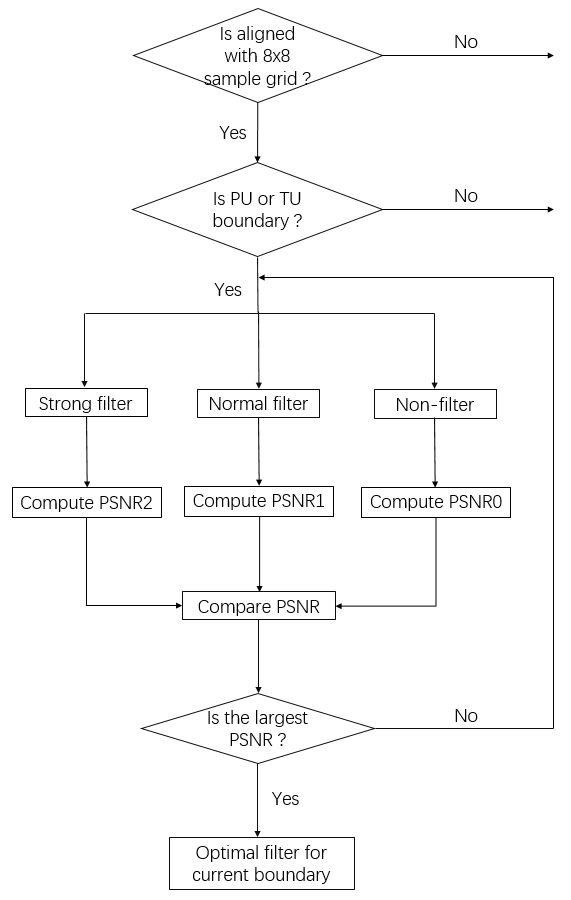}}
 \center{\caption{\label{figureboundary} The proposed filtering decisions for each four-sample segment of the block boundary}} \vspace{-0.03\linewidth}
 \end{figure}
\noindent \textbf{Boundary Experiment} Actualy, to reduce the compression artifacts of block boundary, deblocking filter (DBF) \cite{List20} is adopted as an in-loop filter in HEVC, which applies a set of low-pass filters to 8$\times$8 block boundaries adaptively based on the characteristics of reconstructed samples on both sides of block boundaries, and HEVC only applies DF to samples adjacent to a PU or TU boundary. As a result, more pixels will be changed in the edge or sharping area, and less pixels will be changed in the smooth area. This means that we can train a network to concentrate more on the TU boundary, so that it can do better in the blocking boundary.

To verify this idea, as shown in Figure \ref{figureboundary}, we designed a new filtering mode decision algorithm to test the upper bound of boundary filtering. Specifically, we check all types of filtering modes on both sides of the adjacent blocks, that is, strong filtering mode, normal filtering mode and non-filtering mode, to get an optimal one for each block boundary. Based on the new filtering mode decision, we test 16 widely used video sequences in the common test conditions (CTCs)\cite{CTC}, under AI, RA and LD configurations, respectively. We found that the determined deblocking filtering mode in HEVC is only 35$\%\sim $ 40$\%$ probability to get an optimal one for each block. Therefore, it is possible to get a further improvement on the filtering efficiency. As shown in Table \ref{newmode}, there are 4.2$\%$, 5.9$\%$ and 8.6$\%$ BD-rate saving utilizing the new filtering mode decisions algorithm for AI, RA and LD configurations, respectively.
Through experiments, we found that although the DBF in HEVC can achieve substantial objective and subjective quality improvement, it still can be further improved, which  encourages us to use the boundary information in the decoded bitstream to guide the network to enhance the quality of compressed video.

\renewcommand{\arraystretch}{2.1}
\begin{table}[!htp]
  \centering
  \fontsize{6.5}{6.0}\selectfont
  \caption{Results of the proposed deblocking filter}
  \label{newmode}
    \begin{tabular}{|p{1.5cm}<{\centering}| p{1.5cm}<{\centering}|p{1.0cm}<{\centering}|p{1.0cm}<{\centering}|p{1.0cm}<{\centering}|}
\cline{1-5}
\multicolumn{2}{|c|}{\multirow{1}{*}{{\bf Seqences}}} & \multicolumn{1}{c|}{ { \bf AI }}& \multicolumn{1}{c|}{ { \bf LD }} &\multicolumn{1}{c|}{ { \bf RA }}\cr
\cline{3-5}
\Xhline{0.5pt}
\multicolumn{1}{ |c }{\multirow{5}{*}{{\bf Class B}} } & \multicolumn{1}{ |c| }{Kimono}               & - 3.3$\%$ & - 6.3$\%$ & - 4.2$\%$  \\
\cline{2-5}
\multicolumn{1}{ |c }{}                            & \multicolumn{1}{ |c| }{ParkScene}                & - 5.0$\%$ & - 11.2$\%$ & - 7.1$\%$ \\
\cline{2-5}
\multicolumn{1}{ |c }{}                            & \multicolumn{1}{ |c| }{Cactus}                   & - 5.2$\%$ & - 9.6$\%$ & - 6.4$\%$ \\
\cline{2-5}
\multicolumn{1}{ |c }{}                            & \multicolumn{1}{ |c| }{BasketballDrive}          & - 6.0$\%$ & - 8.9$\%$ & - 6.9$\%$ \\
\cline{2-5}
\multicolumn{1}{ |c }{}                            & \multicolumn{1}{ |c| }{BQTerrace}                & - 3.7$\%$ & - 10.5$\%$ & - 6.5$\%$ \\
\cline{1-5}
\multicolumn{1}{ |c }{\multirow{4}{*}{{\bf Class C}} }    & \multicolumn{1}{ |c| }{BasketballDrill}   & - 4.7$\%$ & - 8.0$\%$ & - 6.2$\%$ \\
\cline{2-5}
\multicolumn{1}{ |c }{}                            & \multicolumn{1}{ |c| }{BQMall}                   & - 4.3$\%$ & - 9.0$\%$ & - 6.3$\%$ \\
\cline{2-5}
\multicolumn{1}{ |c }{}                            & \multicolumn{1}{ |c| }{PartyScene}               & - 3.0$\%$ & - 6.6$\%$ & - 4.8$\%$ \\
\cline{2-5}
\multicolumn{1}{ |c }{}                            & \multicolumn{1}{ |c| }{RaceHorses}               & - 3.4$\%$ & - 7.9$\%$ & - 6.0$\%$ \\
\cline{1-5}
\multicolumn{1}{ |c }{\multirow{4}{*}{{\bf Class D}} }    & \multicolumn{1}{ |c| }{BasketballPass}    & - 4.5$\%$ & - 8.8$\%$ & - 5.6$\%$\\
\cline{2-5}
\multicolumn{1}{ |c }{}                            & \multicolumn{1}{ |c| }{BQSquare}                 & - 2.5$\%$ & - 5.7$\%$ & - 3.4$\%$ \\
\cline{2-5}
\multicolumn{1}{ |c }{}                            & \multicolumn{1}{ |c| }{BlowingBubbles}           & - 3.8$\%$ & - 8.0$\%$ & - 6.0$\%$ \\
\cline{2-5}
\multicolumn{1}{ |c }{}                            & \multicolumn{1}{ |c| }{RaceHorses}               & - 3.6$\%$ & - 9.0$\%$ & - 6.3$\%$\\
\cline{1-5}
\multicolumn{1}{ |c }{\multirow{3}{*}{{\bf Class E}} }    & \multicolumn{1}{ |c| }{FourPeople}        & - 4.9$\%$ & - 9.7$\%$ & - 5.9$\%$ \\
\cline{2-5}
\multicolumn{1}{ |c }{}                            & \multicolumn{1}{ |c| }{Johnny}                   & - 5.1$\%$ & - 9.6$\%$ & - 6.3$\%$ \\
\cline{2-5}
\multicolumn{1}{ |c }{}                            & \multicolumn{1}{ |c| }{KristenAndSara}           & - 4.7$\%$ & - 9.4$\%$ & - 5.8$\%$ \\
\cline{1-5}
\Xhline{0.5pt}
\multicolumn{2}{ |c| }{{\bf Average}} & {\bf - 4.2$\%$} & {\bf - 8.6$\%$}& {\bf - 5.9$\%$}  \\
\Xhline{0.5pt}
    \end{tabular}
\end{table}

Based on the above analysis and findings, in this paper, we fully take advantage of the intra-frame prior information and multi-frame information to design a network with superior performance and robustness for quality enhancement of compressed video.
\section{Network Architecture}
 Our proposed network takes a sequence of    ${N_F} = (2T + 1) \times {{\bf I}^L}$ low quality compressed video frames  ($T$ is the size of temporal span in terms of number of frames), where $\Omega  = \left\{ {{\bf I}_{ - T}^L, \cdots ,{\bf I}_0^L, \cdots ,{\bf I}_T^L} \right\}$ and the guided map of center reference frame  ${\bf F}_0^g$ as inputs, and produces one high quality targt frame ${\bf I}_0^H$ corresponds to center reference frame ${\bf I}_0^L$. The overall architecture of the proposed multi-frame guided attention network, which we call MGANet, is illustrated in Figure \ref{figureover}.
\subsection{BRCLSTM Temporal Encoder}
As we know, the temporal redundancy for video content indicates that there are high correlations among neighboring frames. This correlation appears since the physical characteristics (brightness and color, etc.) are similar among neighboring frames. This is because the neighboring frames are captured within very short time intervals, e.g., about 0.17s for videos with the frame rate of 60 Hz. The background usually does not change in such short time intervals, and only some objects may have few changes in position. It means that most of the low frequency components in successive frames are similar. If we train the network to  directly predict target frame, then the network needs to remember most of the pixel values of the input frame. Therefore, we consider designing a recurrent network to learn the residual information of adjacent frames, so that the target frame can fuse more effective information.

It is known that 3D convolutional Network Recurrent Neural Network (RNN) \cite{Liu21_14}, especially those based on LSTM \cite{Hochreiter22}, Multimodal LSTM \cite{Chuan2} or ConvLSTM \cite{Shi23}, have attracted significant attention in exploiting temporal information. We extend LSTM to work in a bidirectional, residual, and convolutional fashion (we call it BRCLSTM).
\begin{figure}[!htp]
  \centering
  \includegraphics[width=0.48\textwidth]{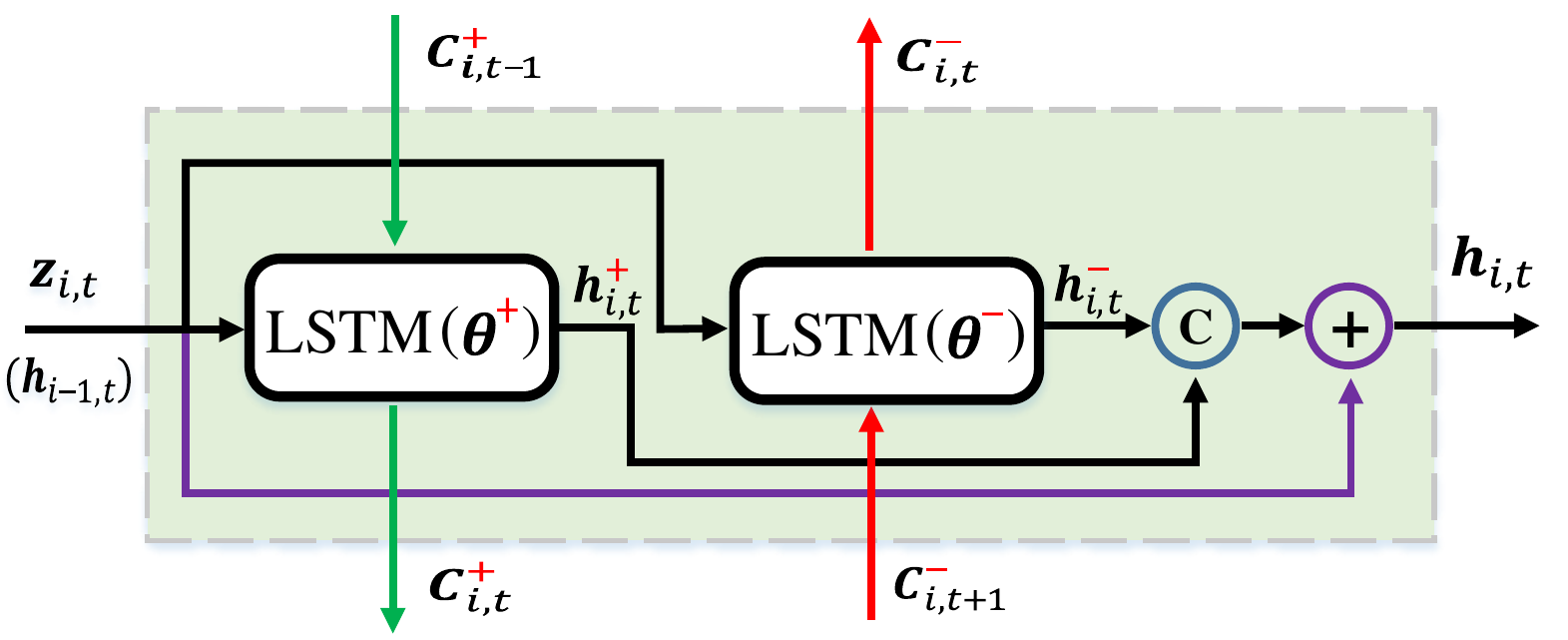}
  \caption{\label{figure5}The proposed BRCLSTM unit}
\end{figure}
\begin{figure*}[htp]
\centering
\includegraphics[width=175mm]  {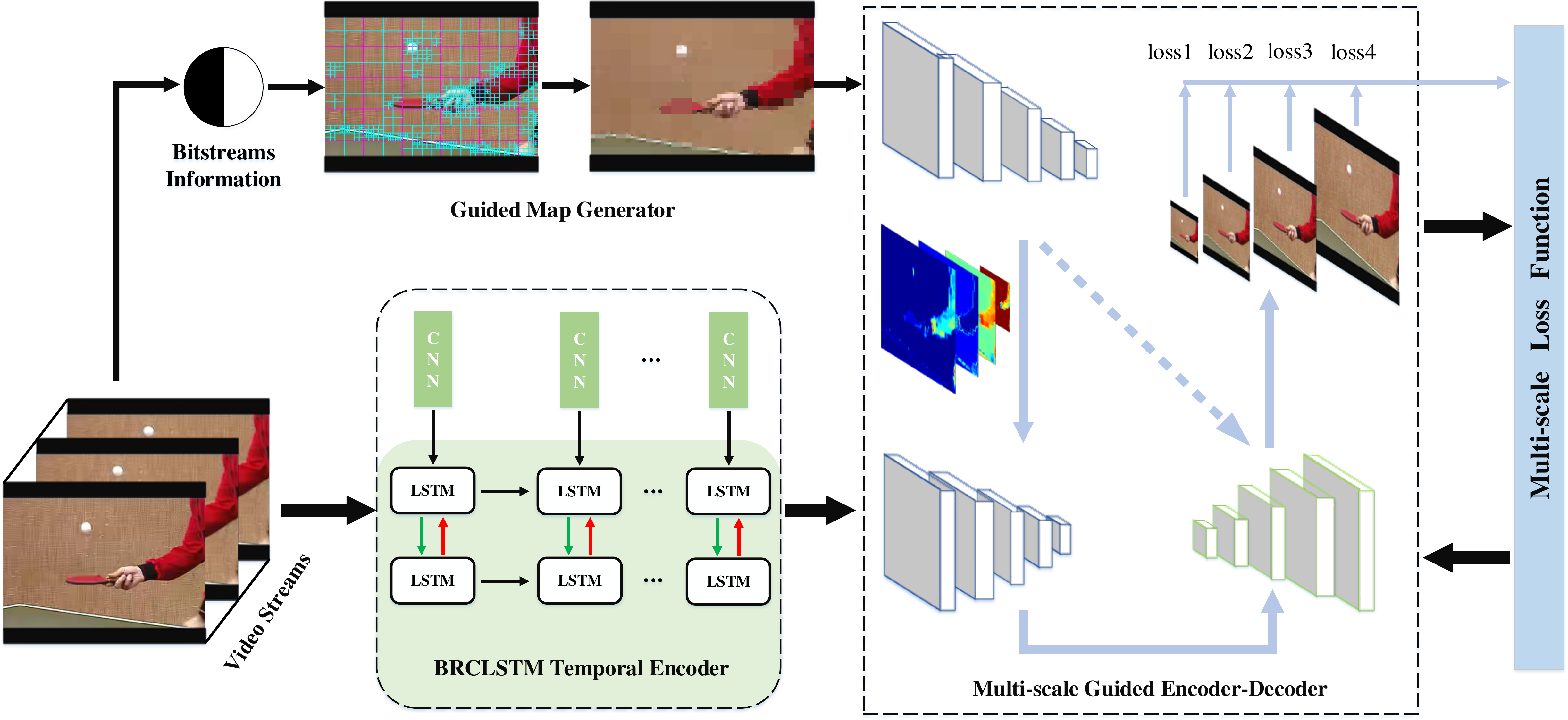}
\caption{\label{figureover}Our proposed MGANet framework} 
\end{figure*}
Specifically, the BRCLSTM temporal encoder consists of a convolutional layer and a BRCLSTM unit for each time step $t$ to implicitly discover frames variations over time, which is used to learn the residual information of adjacent frames.  The convolutional unit is used to  extract the feature maps from the compressed frames, and the extracted feature information is sent to the following BRCLSTM unit. As shown in Figure \ref{figure5}, the structure surrounding by the dotted rectangle is a BRCLSTM unit at the $t$-th frame. Its input, ${\bf{z}}_{i,t}$, is passed to two ConvLSTMs of opposite directions, whose outputs, ${\bf{h}}_{i,t}^ +$ and ${\bf{h}}_{i,t}^ -$, are aggregated and then combined with the input by element-wise add to form the output, ${{\bf{h}}_{i,t}}$. The key equations are shown in equations (\ref{eq3}-\ref{eq6}) below.

\begin{spacing}{0.5}
\begin{equation}\label{eq3}
\left( {\begin{array}{*{20}{c}}
{\bf{F}}\\
{\bf{I}}\\
{\bf{O}}\\
{{\bf{\tilde C}}}
\end{array}} \right) = \left( {\begin{array}{*{20}{c}}
\sigma \\
\sigma \\
\sigma \\
\tau
\end{array}} \right)\left( {{U_i}*{{\bf{Z}}_{i,t}} + {V_i}*{{\bf{H}}_{i,t - 1}} + {{\bf{b}}_i}} \right)
\end{equation}
\end{spacing}
\begin{equation}
{{\bf{C}}_{i,t}} = {\bf{F}} \odot {{\bf{C}}_{i,t - 1}} + {\bf{I}} \odot{{\bf{\tilde C}}}
\end{equation}
\begin{equation}
{{\bf{{\hat h}}}_{i,t}} = {\bf{O}} \odot \tanh ({{\bf{C}}_{i,t}}) \label{eq5}
\end{equation}
\begin{equation}
{{{\bf{ h}}}_{i,t}} = {\cal {C}}({\bf{\hat h}}_{i,t}^ +, {\bf{\hat h}}_{i,t}^ -) \oplus{\bf{z}}_{i,t} \label{eq6}
\end{equation}
where $\odot$ denotes the Hadamard product, $\sigma$ and $\tau$ denote sigmoid and tanh functions, $\cal{C}$ and $*$ denote the concatenation operator and convolution operator, respectively. It worth noting that equations (\ref{eq3}-\ref{eq5}) are the unidirectional expressions only. The BRCLSTM temporal encoder is defined as a multi-layer network, the i-th layer takes the hidden state of the $\left( {i - 1} \right)$-$th$  layer as input: ${{\bf{z}}_{i,t}} = {{\bf{h}}_{i - 1,t}}$, except that the first layer operates on the outputs of the convolutional units, ${{\bf{z}}_{1,t}} = {{\bf{f}}_{1,t}}({\bf{I}}_{1,t}^L)$.
Therefore, the BRCLST temporal encoder can be written as
\begin{equation}
{{\bf{f}}_{i,t}} = {\bf{Ne}}{{\bf{t}}_I}({\bf{I}}_{i,t}^L,{\bf{I}}_{i,t + 1}^L,{\bf{I}}_{i,t - 1}^L,...;{\theta _I}),t \in [ - T,T]
\end{equation}
\begin{equation}
{{\bf{h}}_{i,t}} ={\bf{BRCLSTM}}({{\bf{h}}_{i,t - 1}^+} ,{{\bf{h}}_{i,t + 1}^-},{{\bf{f}}_{i,t}};{ {\theta} _{LSTM}})
\end{equation}
where ${{\bf{Net}}_I}$ and ${\bf{BRCLSTM}}$ are the expressions of convolutional unit and BRCLSTM unit with parameters ${{\theta}}_{I}$ and ${\theta}_{LSTM}$, ${{\theta} _{LSTM}}$ is the set of parameters in BRCLSTM.
\subsection{Guided Encoder-Decoder Network}
The encoder-decoder structure has been proven to be effective in many image/video vision tasks \cite{Unet1,Unet2,Unet3,Unet4,Unet5}.  Particularly, Tao \textit {et al.} \cite{Unet3} proposed an encoder-decoder
resblock network for image deblurring. Wang \textit {et al.} \cite{Chuan1} developed a combined encoder-decoder network of 2DCNN and 3DCNN, which can fill the missing regions inside a video caused by corruption or editing.
In this paper, we design a multi-scale guided encoder-decoder structure with skip-connections. We use the guided map to guide our proposed network to concentrate more on the block boundary. The guided map is fused into our MGANet by a guided attention encoder-decoder subnet.

\begin{figure*}[!htp]
\centering
\includegraphics[width=155mm]  {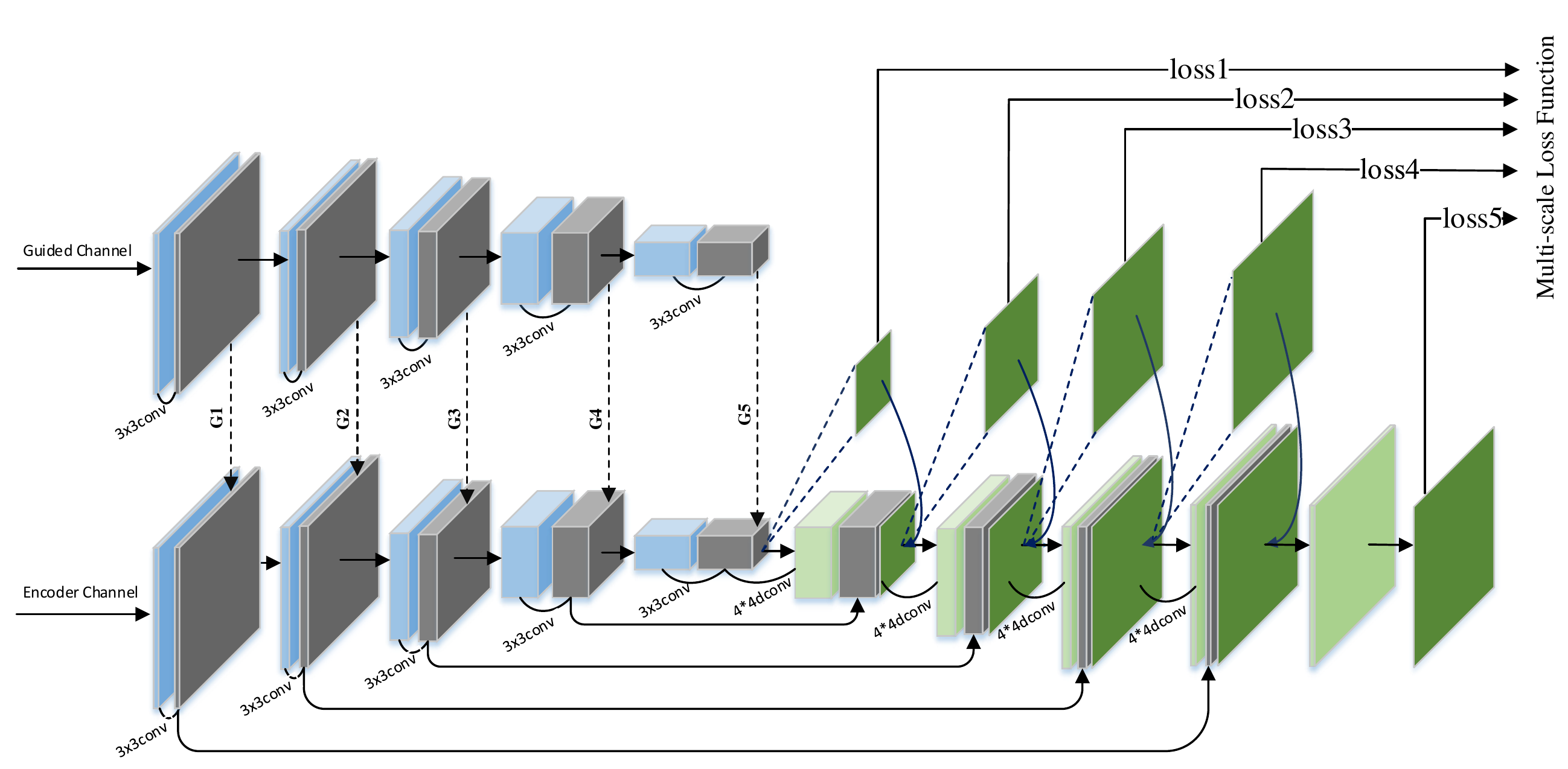}
\caption{\label{figureunet} The proposed guided encoder-decoder subnet. Best viewed in color}
\end{figure*}

\begin{figure}[!htb]
\centering
\subfigure{
\begin{minipage}[t]{0.225\textwidth}
\centering
\includegraphics[width=1.0\textwidth]{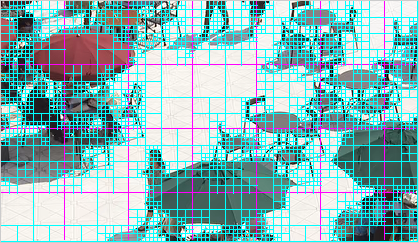}
\end{minipage}
}
\subfigure{
\begin{minipage}[t]{0.225\textwidth}
\centering
\includegraphics[width=1.0\textwidth]{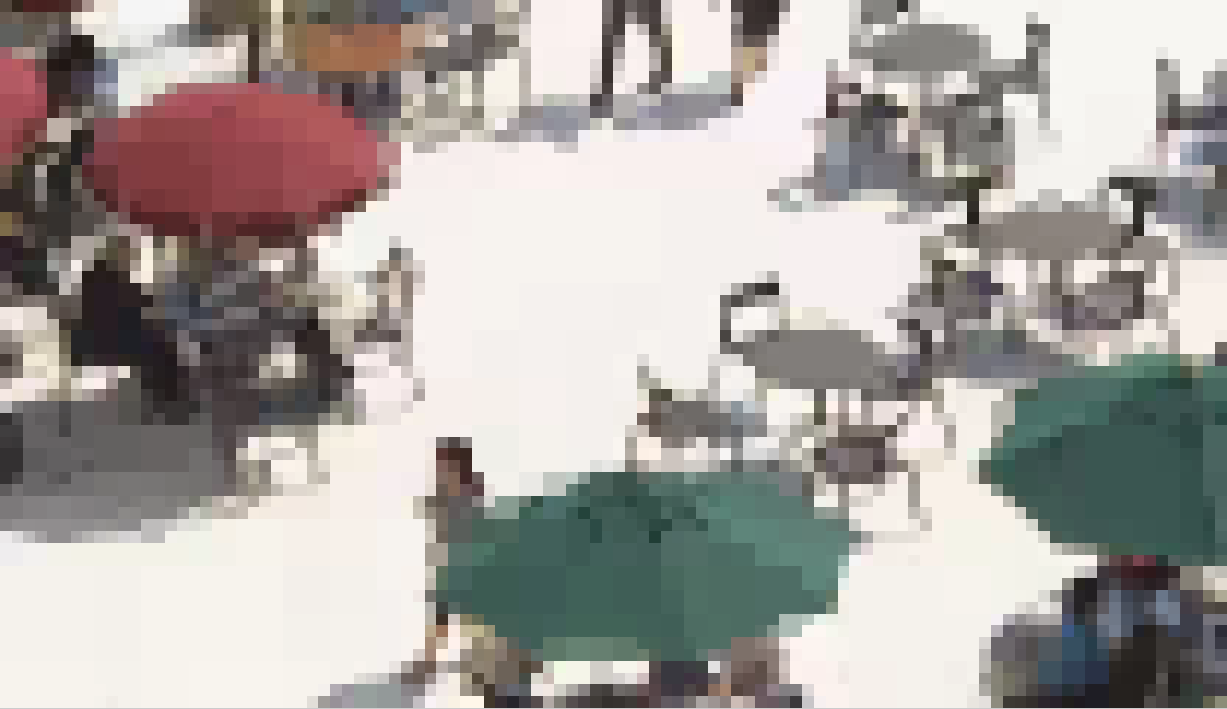}
\end{minipage}
}
\centering
\caption{\label{guided}TUs' partition (left) and guided map (right)}
\end{figure}

\noindent \textbf{Guided Map Generator}
According to the analysis in Sec. \ref{section3}, borrowing TUs' partition information of compressed video, we propose a novel encoder with a guided map as input to guide the encoder-decoder network to concentrate more on the block boundary of transform units.  The  guided map generator is used to produces a series of guided maps $\left\{ {{\bf J}_{t}^m} \right\}$ from compressed frames $\left\{ {{\bf I}_{t}^L} \right\}$ and compressed bit-streams $\left\{ {{\bf I}_{b}^s}\right\}$, which is expressed as
\begin{equation}
{\bf J}_{t}^p = {{\bf{L}}_{BP}}({\bf I}_{t}^L,{{\bf I}_{b \to p}^s};\alpha)
\end{equation}
\begin{equation}
{\bf J}_{t}^g = {{\bf{L}}_{PG}}({{\bf J}_{t}^p} ,{{\bf I}_{p \to g}^s};\beta)
\end{equation}
\begin{equation}
{\bf J}_{t}^m = {{\bf{L}}_{GM}}({\bf I}_{t}^L,{{\bf J}_{t}^g} ,{{\bf I}_{g \to m}^s};\gamma)
\end{equation}
As shown in Figure \ref{guided}, according to the bit-streams information received by decoder, the depth partition information ${{\bf I}_{b \to p}^s}$ of each TU in the compressed frame $\left\{ {{\bf I}_{t}^L} \right\}$ can be extracted.
The basic idea of the proposed method for generate the guided map is first to obtain the TUs'  partition depth information ${\bf J}_{t}^p$,  then determine the boundary ${\bf J}_{t}^g$ of TU according to the depth of different blocks. Finally, the pixel value of each transform unit of the guided map ${\bf J}_{t}^m$ is replaced with the average value of the corresponding  transform block. From the TU's partition of compressed frame in Figure \ref{guided} (left), we can see the TUs' size for smooth area is relatively large, and conversely, it is relatively small for object edges and texture regions. This trend is the same as the distribution of quantization noise analyzed in Sec. \ref{section3}. From Figure \ref{guided}, intuitively, the guided map preserves the overall structure of the target  frame, and also preserves more details in the edge or sharp  areas.
\\
\noindent \textbf{Multi-scale Guided Encoder-Decoder Subnet} The encoder-decoder network \cite{Unet1, Unet2} refers to those symmetric CNN structures that first transform the input data into feature maps with smaller spatial sizes and then transform them back to the shape of the input (decoder). The skip connection between corresponding feature maps in the encoder-decoder is widely used to combine different levels of information.
Typically, the encoder module contains several levels of convolution with strides, and implements the decoder module using a series of deconvolution layers  \cite{Unet1,Unet2,Unet3, Unet4, Unet5}.  

However, directly using the encoder-decoder network is not the best choice for our quality enhancement task. As shown in Figure \ref{figureunet}, we make several modifications to adapt encoder-decoder networks into our framework.
First, we added a guided map encoder channel to guide our network to concentrate more on the block boundary of transform units, which indirectly helps the encoder channel capture useful information and benefits the cross-scale reconstruction.
Second, the two-channel encoders  sharing network weights across scales to significantly reduce training difficulty and introduce obvious stability benefits, which actually amounts to data augmentation. In addition, it reduces the number of trainable parameters significantly.
Finally, to achieve high quality reconstruction  output, we use a multi-supervised loss function to supervise the intermediate output in each scale of the decoder.

It worth noting that the input for each channel is downsampled through four encoder (strided convolution) layers, while the guided encoder channel guides the encoder channel by sum operation at the corresponding convolutional layer, that is, $G_2$, $G_3$, $G_4$ and $G_5$. Meanwhile, the input of encoder channel is also guided by $G_1$. The activations are then passed through four decoder (upsample convolution) layers, with skip connections to the corresponding encoder layer. In addition, each set of decoder activations is passed through another depth-wise convolution layer to generate an intermediate prediction at its resolution. A loss is applied to this intermediate prediction, and the prediction is also concatenated to the decoder activations.

Suppose ${\bf{H}}_{t}^{LSTM  }$ is the final output expression of the BRCLSTM temporal encoder, then,
the modified guided encoder-decoder network can be expressed as
\begin{equation}
{\cal{{\bf F}}}_t^g = {\bf{Ne}}{{\bf{t}}_E}({\bf J}_{t}^{m \downarrow};{\theta _{E}})
\end{equation}
\begin{equation}
{\bf{F}}_t^l = {\bf{Ne}}{{\bf{t}}_E}({\bf{H}}_{t}^{LSTM \downarrow },{\cal{{\bf F}}}_t^g ;{\theta _{E}})
\end{equation}
\begin{equation}
{\bf{F}}_t^d = {\bf{Ne}}{{\bf{t}}_{GD}}({\bf{F}}_t^l,{\cal{{\bf F}}}_t^g;{\theta _{GD}})
\end{equation}
where ${\bf{Ne}}{{\bf{t}}_E}$ and  ${\bf{Ne}}{{\bf{t}}_{GD}}$ are encoder and decoder CNNs with parameters $\theta_E$ and $\theta_{GD}$.
\subsection{Multi-Supervised MGANet} 
The goal of our network is to learn a mapping function $G$ from ${\bf{I}}^L$ to ${\bf{I}}^H$ given training samples $\left\{ {\left( {{\bf{I}}_t^L,{\bf{F}}_0^g} \right)} \right\}_{t = -T}^T$, where, ${\bf F}_0^g$ is the guided map corresponds to center reference frame ${\bf I}_0^L$.
In this work, we consider to minimize the difference between the reconstructed target frame and the ground truth relying on the ${\ell _2}$-loss. We have also tried total variation and adversarial loss, but we notice that ${\ell _2}$-norm is good enough to generate better results for our task. We generate an intermediate prediction of each upsampled block output in the decoder and send it to the loss function, all of the intermediate predictions are supervised during training by loss term  ${{\cal H}_i^{U \uparrow }}$. The loss function of our multi-supervised MGANet can be formulated as,
\begin{equation} \label{eq15}
{ {\cal L}_2} = {\cal L}_0^F +{\lambda _i} \sum\limits_{i = 1}^M {{\cal H}_i^{U \uparrow }}
\end{equation}
where,  ${\cal L}_0^F$ is the final reconstruction loss, and M denotes the number of intermediate predictions.
\begin{equation}
{\lambda _i} = {2^{ - i}},i \in [1,M]
\end{equation}
\begin{equation}
{\cal L}_0^F = \sum\limits_{n = 1}^N {{{\left\| {{G_0}({\bf{I}}_t^L,{\bf{F}}_0^g) - {\bf{I}}_n^H} \right\|}_2}} , t \in [-T,T]
\end{equation}
\begin{equation}
{\cal H}_i^{U \uparrow } = \sum\limits_{n = 1}^N {{{\left\| {G_i^{U \uparrow }({\bf{I}}_t^L,{\bf{F}}_0^g) - {\bf{I}}_n^H} \right\|}_2}}, t \in [-T,T]
\end{equation}
\section{Experiments}
We implement our framework on PyTorch. For fairness, unless noted, otherwise, all experiments are conducted on the same dataset with the same training configuration. Our experiments are conducted on a PC with Intel Xeon E5 CPU and Nvidia GeForce GTX 1080Ti GPU.

\noindent {\bf{Data Preparation}}
For the quality enhancement of compressed video task, training data needs to be of high-quality without noise while containing rich fine details. To achieve good generalization,  similar to \cite{Yang16}, we randomly collect 60  training videos from  the Derf’s collection{\color{red} \footnote{{\color{red} http://media.xiph.org/video/derf/}}}. For the test dataset, 18  sequences of Classes A-E with different resolutions from the Joint Collaborative Team on Video Coding (JCT-VC) standard test set \cite{dataset} are used in our fairness experiments, which are widely used in the development of HEVC standards.  Class A-E is the same as that in the common test conditions (CTCs) \cite{CTC}. This test dataset covers different scene conditions and can better verify the robust of different approaches.

The training and test videos are compressed by the latest HEVC reference software, HM16.9, under All Intra (AI) and Low-Delay (LD) configurations. We set the Quantization Parameters (QPs) to 32, 37 and 42, respectively. When training the models, in each raw clip and its compressed clip, we randomly select the raw frame, its corresponding decoded target frame, and the adjacent frames, together with the guided map to form the training frame pairs. For each frame pair, we divide them into 96$\times$96 sub-images.

\noindent {\bf{Model Training}}
\renewcommand{\arraystretch}{2.2}
\begin{table*}[!htp]
\setlength{\abovecaptionskip}{18pt}
\centering
\fontsize{8.0}{8.5}\selectfont
\caption{Guided Encoder-Decoder Subnet Parameters}
\label{parameter}
\begin{tabular}{m{1.2cm}<{\centering}m{1.2cm}<{\centering}m{1.2cm}<{\centering}m{1.2cm}<{\centering}m{1.2cm}<{\centering}cm{1.2cm}<{\centering}m{1.2cm}<{\centering}m{1.2cm}<{\centering}m{1.2cm}<{\centering}m{1.2cm}<{\centering}}
\hline
\multicolumn{5}{c||}{\multirow{1}{*}{Encoder/Guided (Shared Weights)}}&\multicolumn{5}{c}{\multirow{1}{*}{Guided Decoder}} \\
\hline
\multicolumn{1}{c|}{Layer No.}&Type&Kernel&Stride&\multicolumn{1}{c||}{Channel}&\multicolumn{1}{c|}{Layer No.}&Type&Kernel&Stride&Channel\\
\hline
\multicolumn{1}{c|}{1$^*$}&conv.$\downarrow$&7&2&\multicolumn{1}{c||}{128}&\multicolumn{1}{c|}{1$^\circ $}&deconv&4&2&512\\
\multicolumn{1}{c|}{2}                                    &conv.&3&1&\multicolumn{1}{c||}{128}&\multicolumn{1}{c|}{2}&deconv&4&2&1\\
\multicolumn{1}{c|}{3$^*$}&conv.$\downarrow$&3&2&\multicolumn{1}{c||}{256}&\multicolumn{1}{c|}{3$^\circ $}&deconv&4&2&256\\
\multicolumn{1}{c|}{4}                                   &conv.&3&1&\multicolumn{1}{c||}{256}&\multicolumn{1}{c|}{4}&deconv&4&2&1\\
\multicolumn{1}{c|}{5$^*$}&conv.$\downarrow$&3&2&\multicolumn{1}{c||}{512}&\multicolumn{1}{c|}{5$^\circ $}&deconv&4&2&128\\
\multicolumn{1}{c|}{6}                                    &conv.&3&1&\multicolumn{1}{c||}{512}&\multicolumn{1}{c|}{6}&deconv&4&2&1\\
\multicolumn{1}{c|}{7$^*$}&conv.$\downarrow$&3&2&\multicolumn{1}{c||}{1024}&\multicolumn{1}{c|}{7$^\circ $}&deconv&4&2&64\\
\multicolumn{1}{c|}{8}                                    &conv.&3&1&\multicolumn{1}{c||}{1024}&\multicolumn{1}{c|}{8}&deconv&4&2&1\\
\cline{1-10}
\multicolumn{10}{c}{~}\\[-16pt]
\cline{1-10}
\end{tabular}
\begin{tabular}{m{1.2cm}<{\centering}m{1.2cm}<{\centering}m{1.2cm}<{\centering}m{1.2cm}<{\centering}m{1.2cm}<{\centering}m{0.001cm}<{\centering}m{1.2cm}<{\centering}m{1.2cm}<{\centering}m{1.2cm}<{\centering}m{1.2cm}<{\centering}m{1.2cm}<{\centering}}
\end{tabular}
\end{table*}
All our models are trained following the same protocol and share similar hyperparameters, the detailed parameters for guided encoder-decoder subnet is shown in Table \ref{parameter}. In the table, $*$ represents the layers where downsampling operation and guided operation takes places,  and $\circ $ represents the deconvolutional operation layer corresponding to the guided encoder convolutional layer in the subnet. Filter sizes for convolutional layers are set to 3$\times$3, and all non-linearities  are rectified linear units except for the output layer, which uses a linear activation. Biases are initialized to 0. During training, we use a mini-batch size of 8. To minimize the loss functions of (\ref{eq15}), we employ Adam optimizer \cite{Adam},  start with a learning rate of 1e-4, decay the learning rate with a power of 10 at the $15^{th}$ epochs, and terminate training at 30 epochs. In order to save the training time, we first train the model at QP 42 from scratch and the  models at QP 32 and QP 37  are fine-tuned from it.
\subsection{Quantitative Evaluation} 
To confirm the ability of the proposed network for compressed video, in this section, we evaluate the quality enhancement performance of our MGANet in terms of $\Delta$PSNR, which measures the PSNR difference between the enhanced and the original compressed sequence. We compare our network with some state-of-the-art algorithms, that is, ARCNN \cite{Dong9_15}, VRCNN \cite{Dai14}, MemNet \cite{Tai17}, DnCNN \cite{Zhang13_17}, DCAD \cite{Wang15_17Chao} and MFQE \cite{Yang16} for compressed video. Among them, DnCNN and MemNet are the latest quality enhancement approaches for compressed image, MFQE is the state-of-the-art video quality enhancement approach. 
\renewcommand{\arraystretch}{1.8}
\begin{table}[!htp]
\centering
\begin{threeparttable}
\setlength{\abovecaptionskip}{8pt}
\centering
\fontsize{5.4}{6.5}\selectfont
\caption{Overall $\Delta$PSNR (dB) of the test sequences under LD configuration, red color
indicates the best performance and blue color indicates the second best performance.}
\label{tab:table1}
\begin{tabular}{|m{0.15cm}<{\centering}|m{0.15cm}<{\centering}|m{0.6cm}<{\centering}|m{0.6cm}<{\centering}|m{0.6cm}<{\centering}|m{0.6cm}<{\centering}|m{0.6cm}<{\centering}|m{0.6cm}<{\centering}|m{0.6cm}<{\centering}|}
\cline{1-8}
 \multicolumn{1}{|c}{\multirow{1}{*}{{\bf Class}}}
&\multicolumn{1}{|c|}{\multirow{1}{*}{{\bf Seq}}}
&\multicolumn{1}{c|}{\makecell{\bf ARCNN\\ \bf \cite{Dong9_15}}}
&\multicolumn{1}{c|}{\makecell{\bf MemNet\\ \bf \cite{Tai17}}}
&\multicolumn{1}{c|}{\makecell{\bf DnCNN\\ \bf \cite{Zhang13_17}}}
&\multicolumn{1}{c|}{\makecell{\bf DCAD\\ \bf \cite{Wang15_17Chao}}}
&\multicolumn{1}{c|}{\makecell{\bf MFQE\\ \bf \cite{Yang16}}}
&\multicolumn{1}{c|}{\makecell{\bf MGANet\\ \bf(ours)}}\cr
\cline{1-8}
 \multicolumn{1}{|c }{\multirow{2}{*}{{\bf A}}}
&\multicolumn{1}{|c|}{1}
&\multicolumn{1}{c|}{0.4637}
&\multicolumn{1}{c|}{0.4841}
&\multicolumn{1}{c|}{0.3980}
&\multicolumn{1}{c|}{0.2791}
&\multicolumn{1}{c|}{\color{blue}{\bf{0.7026}}}%
&\multicolumn{1}{c|}{{\color{red}{\bf{0.7242}}}}\\
\cline{2-8}
\multicolumn{1}{|c }{\multirow{1}{*}{}}
&\multicolumn{1}{|c|}{2}
&\multicolumn{1}{c|}{0.2679}
&\multicolumn{1}{c|}{0.2295}
&\multicolumn{1}{c|}{0.2501}
&\multicolumn{1}{c|}{0.1743}
&\multicolumn{1}{c|}{\color{blue}{\bf{0.2864}}}%
&\multicolumn{1}{c|}{{\color{red}{\bf{0.4602}}}}\\
\cline{1-8}
\multicolumn{1}{|c }{\multirow{5}{*}{{\bf B}}}
&\multicolumn{1}{|c|}{3}
&\multicolumn{1}{c|}{0.2460}
&\multicolumn{1}{c|}{0.2557}
&\multicolumn{1}{c|}{0.2319}
&\multicolumn{1}{c|}{0.1999}
&\multicolumn{1}{c|}{\color{red}{\bf{0.4921}}}%
&\multicolumn{1}{c|}{{\color{blue}{\bf{0.4729}}}}\\
\cline{2-8}
\multicolumn{1}{|c }{\multirow{1}{*}{}}
&\multicolumn{1}{|c|}{4}
&\multicolumn{1}{c|}{0.1691}
&\multicolumn{1}{c|}{0.1754}
&\multicolumn{1}{c|}{0.1550}
&\multicolumn{1}{c|}{0.1253}
&\multicolumn{1}{c|}{\color{red}{\bf{0.2404}}}%
&\multicolumn{1}{c|}{{\color{blue}{\bf{0.2347}}}}\\
\cline{2-8}
\multicolumn{1}{|c }{\multirow{1}{*}{}}
&\multicolumn{1}{|c|}{5}
&\multicolumn{1}{c|}{0.1053}
&\multicolumn{1}{c|}{0.1486}
&\multicolumn{1}{c|}{0.1860}
&\multicolumn{1}{c|}{0.1318}
&\multicolumn{1}{c|}{\color{blue}{\bf{0.2676}}}%
&\multicolumn{1}{c|}{{\color{red}{\bf{0.3719}}}}\\
\cline{2-8}
\multicolumn{1}{|c }{\multirow{1}{*}{}}
&\multicolumn{1}{|c|}{6}
&\multicolumn{1}{c|}{0.1803}
&\multicolumn{1}{c|}{0.2173}
&\multicolumn{1}{c|}{0.2135}
&\multicolumn{1}{c|}{0.1077}
&\multicolumn{1}{c|}{\color{blue}{\bf{0.2189}}}%
&\multicolumn{1}{c|}{{\color{red}{\bf{0.3251}}}}\\
\cline{2-8}
\multicolumn{1}{|c }{\multirow{1}{*}{}}
&\multicolumn{1}{|c|}{7}
&\multicolumn{1}{c|}{0.1102}
&\multicolumn{1}{c|}{\color{blue}{\bf{0.1717}}}
&\multicolumn{1}{c|}{0.1008}
&\multicolumn{1}{c|}{-0.0755}
&\multicolumn{1}{c|}{-0.1132}%
&\multicolumn{1}{c|}{{\color{red}{\bf{0.1841}}}}\\
\cline{1-8}
\multicolumn{1}{|c }{\multirow{4}{*}{{\bf C}}}
&\multicolumn{1}{|c|}{8}
&\multicolumn{1}{c|}{0.1559}
&\multicolumn{1}{c|}{0.1502}
&\multicolumn{1}{c|}{0.1055}
&\multicolumn{1}{c|}{0.0529}
&\multicolumn{1}{c|}{\color{blue}{\bf{0.1766}}}%
&\multicolumn{1}{c|}{{\color{red}{\bf{0.4159}}}}\\
\cline{2-8}
\multicolumn{1}{|c }{\multirow{1}{*}{}}
&\multicolumn{1}{|c|}{9}
&\multicolumn{1}{c|}{0.1667}
&\multicolumn{1}{c|}{\color{blue}{\bf{0.2270}}}
&\multicolumn{1}{c|}{0.1421}
&\multicolumn{1}{c|}{0.0641}
&\multicolumn{1}{c|}{0.0725}%
&\multicolumn{1}{c|}{{\color{red}{\bf{0.3983}}}}\\
\cline{2-8}
\multicolumn{1}{|c }{\multirow{1}{*}{}}
&\multicolumn{1}{|c|}{10}
&\multicolumn{1}{c|}{0.0224}
&\multicolumn{1}{c|}{\color{blue}{\bf{0.0961}}}
&\multicolumn{1}{c|}{0.0052}
&\multicolumn{1}{c|}{0.0511}
&\multicolumn{1}{c|}{-0.1589}%
&\multicolumn{1}{c|}{{\color{red}{\bf{0.2519}}}}\\
\cline{2-8}
\multicolumn{1}{|c }{\multirow{1}{*}{}}
&\multicolumn{1}{|c|}{11}
&\multicolumn{1}{c|}{0.1442}
&\multicolumn{1}{c|}{\color{blue}{\bf{0.1732}}}
&\multicolumn{1}{c|}{0.1178}
&\multicolumn{1}{c|}{0.0770}
&\multicolumn{1}{c|}{0.0052}%
&\multicolumn{1}{c|}{{\color{red}{\bf{0.1868}}}}\\
\cline{1-8}
\multicolumn{1}{|c }{\multirow{4}{*}{{\bf D}}}
&\multicolumn{1}{|c|}{12}
&\multicolumn{1}{c|}{0.1927}
&\multicolumn{1}{c|}{0.1892}
&\multicolumn{1}{c|}{0.1487}
&\multicolumn{1}{c|}{0.1180}
&\multicolumn{1}{c|}{\color{blue}{\bf{0.3936}}}%
&\multicolumn{1}{c|}{{\color{red}{\bf{0.4865}}}}\\
\cline{2-8}
\multicolumn{1}{|c }{\multirow{1}{*}{}}
&\multicolumn{1}{|c|}{13}
&\multicolumn{1}{c|}{-0.1108}
&\multicolumn{1}{c|}{\color{blue}{\bf{0.0508}}}
&\multicolumn{1}{c|}{-0.0806}
&\multicolumn{1}{c|}{-0.1223}
&\multicolumn{1}{c|}{-0.4418}%
&\multicolumn{1}{c|}{{\color{red}{\bf{0.2786}}}}\\
\cline{2-8}
\multicolumn{1}{|c }{\multirow{1}{*}{}}
&\multicolumn{1}{|c|}{14}
&\multicolumn{1}{c|}{0.0933}
&\multicolumn{1}{c|}{0.1347}
&\multicolumn{1}{c|}{0.1701}
&\multicolumn{1}{c|}{0.0350}
&\multicolumn{1}{c|}{\color{blue}{\bf{0.1426}}}%
&\multicolumn{1}{c|}{{\color{red}{\bf{0.2803}}}}\\
\cline{2-8}
\multicolumn{1}{|c }{\multirow{1}{*}{}}
&\multicolumn{1}{|c|}{15}
&\multicolumn{1}{c|}{0.2572}
&\multicolumn{1}{c|}{0.2916}
&\multicolumn{1}{c|}{0.2275}
&\multicolumn{1}{c|}{0.1665}
&\multicolumn{1}{c|}{\color{red}{\bf{0.3861}}}%
&\multicolumn{1}{c|}{{\color{blue}{\bf{0.3179}}}}\\
\cline{1-8}
\multicolumn{1}{|c }{\multirow{3}{*}{{\bf E}}}
&\multicolumn{1}{|c|}{16}
&\multicolumn{1}{c|}{0.4020}
&\multicolumn{1}{c|}{0.3893}
&\multicolumn{1}{c|}{0.3747}
&\multicolumn{1}{c|}{0.2613}
&\multicolumn{1}{c|}{\color{blue}{\bf{0.4997}}}%
&\multicolumn{1}{c|}{{\color{red}{\bf{0.6554}}}}\\
\cline{2-8}
\multicolumn{1}{|c }{\multirow{1}{*}{}}
&\multicolumn{1}{|c|}{17}
&\multicolumn{1}{c|}{0.2332}
&\multicolumn{1}{c|}{0.3692}
&\multicolumn{1}{c|}{0.2665}
&\multicolumn{1}{c|}{0.1602}
&\multicolumn{1}{c|}{\color{blue}{\bf{0.3823}}}%
&\multicolumn{1}{c|}{{\color{red}{\bf{0.5727}}}}\\
\cline{2-8}
\multicolumn{1}{|c }{\multirow{1}{*}{}}
&\multicolumn{1}{|c|}{18}
&\multicolumn{1}{c|}{0.3934}
&\multicolumn{1}{c|}{0.3525}
&\multicolumn{1}{c|}{0.3733}
&\multicolumn{1}{c|}{0.2792}
&\multicolumn{1}{c|}{\color{blue}{\bf{0.4784}}}%
&\multicolumn{1}{c|}{{\color{red}{\bf{0.6561}}}}\\
\cline{1-8}
\multicolumn{2}{|c| }{\multirow{1}{*}{{\bf{QP37 AVE.}}}}
&\multicolumn{1}{c|}{0.1940}
&\multicolumn{1}{c|}{0.2281}
&\multicolumn{1}{c|}{0.1881}
&\multicolumn{1}{c|}{0.1189}
&\multicolumn{1}{c|}{\color{blue}{\bf{0.2545}}}
&\multicolumn{1}{c|}{{\color{red}{\bf{0.4041}}}}\\
\cline{1-8}
\multicolumn{8}{c}{~}\\[-6pt]
\cline{1-8}
\multicolumn{1}{|c }{\multirow{1}{*}{\bf{QP32}}} & \multicolumn{1}{|c|}{\bf{AVE.}} & \multicolumn{1}{c|}{0.1211} &\multicolumn{1}{c|}{\color{blue}{\bf{0.1583}}}&\multicolumn{1}{c|}{0.1104} &\multicolumn{1}{c|}{0.0826} &\multicolumn{1}{c|}{-}&\multicolumn{1}{c|}{\color{red}{\bf{0.3528}}}\\
\cline{1-8}
\multicolumn{8}{c}{~}\\[-11pt]
\cline{1-8}
\multicolumn{1}{|c }{\multirow{1}{*}{\bf{QP42}}} & \multicolumn{1}{|c|}{\bf{AVE.}} & \multicolumn{1}{c|}{0.1327} &\multicolumn{1}{c|}{\color{blue}{\bf{0.1611}}}&\multicolumn{1}{c|}{0.1259} &\multicolumn{1}{c|}{0.0941} &\multicolumn{1}{c|}{-}&\multicolumn{1}{c|}{\color{red}{\bf{0.3463}}}\\
\cline{1-8}
\end{tabular}
\begin{tablenotes}
\item {\bf{Seq}}  1:PeopleOnStreet 2:Traffic 3:Kimono 4:ParkScene 5:Cactus 6:BasketballDrive 7:BQTerrace 8:BasketballDrill 9:BQMall 10:PartyScene 11:RaceHorsesC 12:BasketballPass 13:BQSquare 14:BlowingBubbles 15:RaceHorses 16:FourPeople 17:Johnny 18:KristenAndSara
\end{tablenotes}
\end{threeparttable} 
\end{table}
\renewcommand{\arraystretch}{1.8}
\begin{table}[!htp]
\centering
\begin{threeparttable}
\setlength{\abovecaptionskip}{8pt}
\centering
\fontsize{5.4}{6.5}\selectfont
\caption{Overall $\Delta$PSNR (dB) of the test sequences under AI configuration, red color
indicates the best performance and blue color indicates the second best performance.}
\label{tab:table2}
\begin{tabular}{|m{0.15cm}<{\centering}|m{0.15cm}<{\centering}|m{0.6cm}<{\centering}|m{0.6cm}<{\centering}|m{0.6cm}<{\centering}|m{0.6cm}<{\centering}|m{0.6cm}<{\centering}|m{0.6cm}<{\centering}|m{0.6cm}<{\centering}|}
\cline{1-8}
 \multicolumn{1}{|c}{\multirow{1}{*}{{\bf Class}}}
&\multicolumn{1}{|c|}{\multirow{1}{*}{{\bf Seq}}}
&\multicolumn{1}{c|}{\makecell{\bf ARCNN\\ \bf \cite{Dong9_15}}}
&\multicolumn{1}{c|}{\makecell{\bf DCAD\\ \bf \cite{Wang15_17Chao}}}
&\multicolumn{1}{c|}{\makecell{\bf DnCNN\\ \bf \cite{Zhang13_17}}}
&\multicolumn{1}{c|}{\makecell{\bf VRCNN\\ \bf \cite{Dai14}}}
&\multicolumn{1}{c|}{\makecell{\bf MemNet\\ \bf \cite{Tai17}}}
&\multicolumn{1}{c|}{\makecell{\bf MGANet\\ \bf(ours)}}\cr
\cline{1-8}
 \multicolumn{1}{|c }{\multirow{2}{*}{{\bf A}}}
&\multicolumn{1}{|c|}{1}
&\multicolumn{1}{c|}{0.5220}
&\multicolumn{1}{c|}{0.3667}
&\multicolumn{1}{c|}{0.4475}
&\multicolumn{1}{c|}{0.5673}%
&\multicolumn{1}{c|}{\color{blue}{\bf{0.5756}}}
&\multicolumn{1}{c|}{{\color{red}{\bf{1.0496}}}}\\
\cline{2-8}
\multicolumn{1}{|c }{\multirow{1}{*}{}}
&\multicolumn{1}{|c|}{2}
&\multicolumn{1}{c|}{0.3784}
&\multicolumn{1}{c|}{0.2489}
&\multicolumn{1}{c|}{0.3007}
&\multicolumn{1}{c|}{0.3147}%
&\multicolumn{1}{c|}{\color{blue}{\bf{0.4357}}}
&\multicolumn{1}{c|}{{\color{red}{\bf{1.3862}}}}\\
\cline{1-8}
\multicolumn{1}{|c }{\multirow{5}{*}{{\bf B}}}
&\multicolumn{1}{|c|}{3}
&\multicolumn{1}{c|}{0.2286}
&\multicolumn{1}{c|}{0.1448}
&\multicolumn{1}{c|}{0.1880}
&\multicolumn{1}{c|}{0.2445}%
&\multicolumn{1}{c|}{\color{blue}{\bf{0.2714}}}
&\multicolumn{1}{c|}{{\color{red}{\bf{0.6506}}}}\\
\cline{2-8}
\multicolumn{1}{|c }{\multirow{1}{*}{}}
&\multicolumn{1}{|c|}{4}
&\multicolumn{1}{c|}{0.2503}
&\multicolumn{1}{c|}{0.1708}
&\multicolumn{1}{c|}{0.2092}
&\multicolumn{1}{c|}{0.2634}%
&\multicolumn{1}{c|}{\color{blue}{\bf{0.3520}}}
&\multicolumn{1}{c|}{{\color{red}{\bf{1.0211}}}}\\
\cline{2-8}
\multicolumn{1}{|c }{\multirow{1}{*}{}}
&\multicolumn{1}{|c|}{5}
&\multicolumn{1}{c|}{0.2578}
&\multicolumn{1}{c|}{0.1477}
&\multicolumn{1}{c|}{0.1731}
&\multicolumn{1}{c|}{\color{blue}{\bf{0.3511}}}%
&\multicolumn{1}{c|}{0.2352}
&\multicolumn{1}{c|}{{\color{red}{\bf{0.9401}}}}\\
\cline{2-8}
\multicolumn{1}{|c }{\multirow{1}{*}{}}
&\multicolumn{1}{|c|}{6}
&\multicolumn{1}{c|}{0.1127}
&\multicolumn{1}{c|}{0.0440}
&\multicolumn{1}{c|}{0.0591}
&\multicolumn{1}{c|}{0.1506}%
&\multicolumn{1}{c|}{\color{blue}{\bf{0.1876}}}
&\multicolumn{1}{c|}{{\color{red}{\bf{0.4307}}}}\\
\cline{2-8}
\multicolumn{1}{|c }{\multirow{1}{*}{}}
&\multicolumn{1}{|c|}{7}
&\multicolumn{1}{c|}{0.1598}
&\multicolumn{1}{c|}{0.1386}
&\multicolumn{1}{c|}{0.1531}
&\multicolumn{1}{c|}{0.2139}%
&\multicolumn{1}{c|}{\color{blue}{\bf{0.2456}}}
&\multicolumn{1}{c|}{{\color{red}{\bf{0.8940}}}}\\
\cline{1-8}
\multicolumn{1}{|c }{\multirow{4}{*}{{\bf C}}}
&\multicolumn{1}{|c|}{8}
&\multicolumn{1}{c|}{0.2491}
&\multicolumn{1}{c|}{0.1522}
&\multicolumn{1}{c|}{0.1559}
&\multicolumn{1}{c|}{\color{blue}{\bf{0.3062}}}%
&\multicolumn{1}{c|}{0.2053}
&\multicolumn{1}{c|}{{\color{red}{\bf{1.0267}}}}\\
\cline{2-8}
\multicolumn{1}{|c }{\multirow{1}{*}{}}
&\multicolumn{1}{|c|}{9}
&\multicolumn{1}{c|}{0.1258}
&\multicolumn{1}{c|}{0.0043}
&\multicolumn{1}{c|}{0.0451}
&\multicolumn{1}{c|}{\color{blue}{\bf{0.2052}}}%
&\multicolumn{1}{c|}{0.1241}
&\multicolumn{1}{c|}{{\color{red}{\bf{0.8488}}}}\\
\cline{2-8}
\multicolumn{1}{|c }{\multirow{1}{*}{}}
&\multicolumn{1}{|c|}{10}
&\multicolumn{1}{c|}{0.0284}
&\multicolumn{1}{c|}{0.0808}
&\multicolumn{1}{c|}{0.0951}
&\multicolumn{1}{c|}{0.1450}%
&\multicolumn{1}{c|}{\color{blue}{\bf{0.1521}}}
&\multicolumn{1}{c|}{{\color{red}{\bf{1.1859}}}}\\
\cline{2-8}
\multicolumn{1}{|c }{\multirow{1}{*}{}}
&\multicolumn{1}{|c|}{11}
&\multicolumn{1}{c|}{0.1605}
&\multicolumn{1}{c|}{0.0502}
&\multicolumn{1}{c|}{0.0472}
&\multicolumn{1}{c|}{\color{blue}{\bf{0.1854}}}%
&\multicolumn{1}{c|}{0.1156}
&\multicolumn{1}{c|}{{\color{red}{\bf{0.1903}}}}\\
\cline{1-8}
\multicolumn{1}{|c }{\multirow{4}{*}{{\bf D}}}
&\multicolumn{1}{|c|}{12}
&\multicolumn{1}{c|}{0.1480}
&\multicolumn{1}{c|}{0.0459}
&\multicolumn{1}{c|}{0.0156}
&\multicolumn{1}{c|}{\color{blue}{\bf{0.2596}}}%
&\multicolumn{1}{c|}{0.1623}
&\multicolumn{1}{c|}{{\color{red}{\bf{0.6086}}}}\\
\cline{2-8}
\multicolumn{1}{|c }{\multirow{1}{*}{}}
&\multicolumn{1}{|c|}{13}
&\multicolumn{1}{c|}{0.0331}
&\multicolumn{1}{c|}{0.0548}
&\multicolumn{1}{c|}{0.1157}
&\multicolumn{1}{c|}{\color{blue}{\bf{0.2196}}}%
&\multicolumn{1}{c|}{0.1325}
&\multicolumn{1}{c|}{{\color{red}{\bf{1.4436}}}}\\
\cline{2-8}
\multicolumn{1}{|c }{\multirow{1}{*}{}}
&\multicolumn{1}{|c|}{14}
&\multicolumn{1}{c|}{0.1026}
&\multicolumn{1}{c|}{0.0103}
&\multicolumn{1}{c|}{0.0007}
&\multicolumn{1}{c|}{0.1465}%
&\multicolumn{1}{c|}{\color{blue}{\bf{0.1626}}}
&\multicolumn{1}{c|}{{\color{red}{\bf{1.1067}}}}\\
\cline{2-8}
\multicolumn{1}{|c }{\multirow{1}{*}{}}
&\multicolumn{1}{|c|}{15}
&\multicolumn{1}{c|}{0.3148}
&\multicolumn{1}{c|}{0.2072}
&\multicolumn{1}{c|}{0.2475}
&\multicolumn{1}{c|}{0.3446}%
&\multicolumn{1}{c|}{\color{blue}{\bf{0.4201}}}
&\multicolumn{1}{c|}{{\color{red}{\bf{0.4951}}}}\\
\cline{1-8}
\multicolumn{1}{|c }{\multirow{3}{*}{{\bf E}}}
&\multicolumn{1}{|c|}{16}
&\multicolumn{1}{c|}{0.3324}
&\multicolumn{1}{c|}{0.3197}
&\multicolumn{1}{c|}{0.3845}
&\multicolumn{1}{c|}{0.2667}%
&\multicolumn{1}{c|}{\color{blue}{\bf{0.4523}}}
&\multicolumn{1}{c|}{{\color{red}{\bf{1.8187}}}}\\
\cline{2-8}
\multicolumn{1}{|c }{\multirow{1}{*}{}}
&\multicolumn{1}{|c|}{17}
&\multicolumn{1}{c|}{0.2362}
&\multicolumn{1}{c|}{0.1366}
&\multicolumn{1}{c|}{0.1512}
&\multicolumn{1}{c|}{0.2149}%
&\multicolumn{1}{c|}{\color{blue}{\bf{0.2625}}}
&\multicolumn{1}{c|}{{\color{red}{\bf{1.4235}}}}\\
\cline{2-8}
\multicolumn{1}{|c }{\multirow{1}{*}{}}
&\multicolumn{1}{|c|}{18}
&\multicolumn{1}{c|}{0.3971}
&\multicolumn{1}{c|}{0.2958}
&\multicolumn{1}{c|}{0.3224}
&\multicolumn{1}{c|}{0.3237}%
&\multicolumn{1}{c|}{\color{blue}{\bf{0.4701}}}
&\multicolumn{1}{c|}{{\color{red}{\bf{1.6570}}}}\\
\cline{1-8}
\multicolumn{2}{|c| }{\multirow{1}{*}{{\bf{QP37 AVE.}}}}
&\multicolumn{1}{c|}{0.2237}
&\multicolumn{1}{c|}{0.1203}
&\multicolumn{1}{c|}{0.1404}
&\multicolumn{1}{c|}{0.2339}
&\multicolumn{1}{c|}{\color{blue}{\bf{0.2757}}}
&\multicolumn{1}{c|}{{\color{red}{\bf{1.0049}}}}\\
\cline{1-8}
\multicolumn{8}{c}{~}\\[-6pt]
\cline{1-8}
\multicolumn{1}{|c }{\multirow{1}{*}{\bf{QP32}}} & \multicolumn{1}{|c|}{\bf{AVE.}} & \multicolumn{1}{c|}{0.1822} &\multicolumn{1}{c|}{0.1164}&\multicolumn{1}{c|}{0.1355} &\multicolumn{1}{c|}{0.1973} &\multicolumn{1}{c|}{\color{blue}{\bf{0.2305}}}&\multicolumn{1}{c|}{\color{red}{\bf{0.8387}}}\\
\cline{1-8}
\multicolumn{8}{c}{~}\\[-11pt]
\cline{1-8}
\multicolumn{1}{|c }{\multirow{1}{*}{\bf{QP42}}} & \multicolumn{1}{|c|}{\bf{AVE.}} & \multicolumn{1}{c|}{0.2031} &\multicolumn{1}{c|}{0.1442}&\multicolumn{1}{c|}{0.1570} &\multicolumn{1}{c|}{0.2082} &\multicolumn{1}{c|}{\color{blue}{\bf{0.2547}}}&\multicolumn{1}{c|}{\color{red}{\bf{0.9998}}}\\
\cline{1-8}
\end{tabular}
\begin{tablenotes}
\item {\bf{Seq}}  1:PeopleOnStreet 2:Traffic 3:Kimono 4:ParkScene 5:Cactus 6:BasketballDrive 7:BQTerrace 8:BasketballDrill 9:BQMall 10:PartyScene 11:RaceHorsesC 12:BasketballPass 13:BQSquare 14:BlowingBubbles 15:RaceHorses 16:FourPeople 17:Johnny 18:KristenAndSara
\end{tablenotes}
\end{threeparttable}
\end{table}

For MFQE{\color{red} \footnote{{\color{red} https://github.com/ryangBUAA/MFQE}}} approach and VRCNN{\color{red} \footnote{\color{red} https://github.com/dongeliu/ilfcnn}} approach, we run the test code provided by authors directly and make a fair comparison with our method. Our source codes and the database for the TU partition of HEVC are available at Github \url{https://github.com/mengab/MGANet} . Since MFQE approach just has the test model at QP 37 for LD configuration, we only test the MFQE model at QP 37 in the experiment. VRCNN just has the test model under AI configuration at QP 32 and QP 37, we retrained the existing networks utilizing author's training code with the recommended parameters for the AI configuration on \textit{Caffe}  \cite{caffe} at QP 42.  Other existing networks also use the same training dataset and authors' recommended parameters to retrain on PyTorch. We randomly test consecutive 20 frames of each test sequence and then averaged them over all the frames as the final result for all the models.

Table \ref{tab:table1} and Table \ref{tab:table2} present the $\Delta$PSNR results of  the test sequences under AI and LD configurations, respectively.
In overall, our MGANet approach outperforms all other compared approaches for the test sequences on average.
To be specific, for LD configuration, the highest $\Delta$PSNR of our MGANet reaches 0.7242dB at QP 37, and the averaged $\Delta$PSNR of our MGANet approach is 0.4041 dB, it is much higher than that of MFQE approach \cite{Yang16} (0.2545 dB), which is state-of-the-art in the compared methods.
Our model  is more robust than the comparison methods, especially for the sequences `BQTerrace',  `PartyScene'  and `BQSquare'.

In addition, our MGANet approach significantly outperforms all other caparison approaches under AI configuration. As shown in Table \ref{tab:table2}, the highest $\Delta$PSNR of our MGANet reaches 1.8187dB at QP 37, the averaged $\Delta$PSNR of our MGANet approach is 1.0049 dB, which is much better than that of MemNet (0.2757dB), DCAD (0.1203dB), DnCNN (0.1404dB) and VRCNN (0.2339dB). Among them, the MemNet is much deeper than our MGANet, which is stacked more than 80 convolutional layers to reconstruct the artifact images. Thus, our MGANet approach is effective in the quality enhancement of compressed video.
We can also observe that our network achieves a higher coding gain than the LD configuration in the AI configuration. LD employs inter-prediction and complex hierarchical frame structure, which makes the residue has a lower correlation. In such a case, the training set has lower quality samples and this results in a lower coding gain.

From Tables \ref{tab:table1} and  \ref{tab:table2}, we can infer that image prior information and multi-frame information play an important role in the quality enhancement of compressed video. 
\subsection{Quality Fluctuation} 
\begin{figure}[htb]
\begin{center}
\includegraphics[width=85mm]  {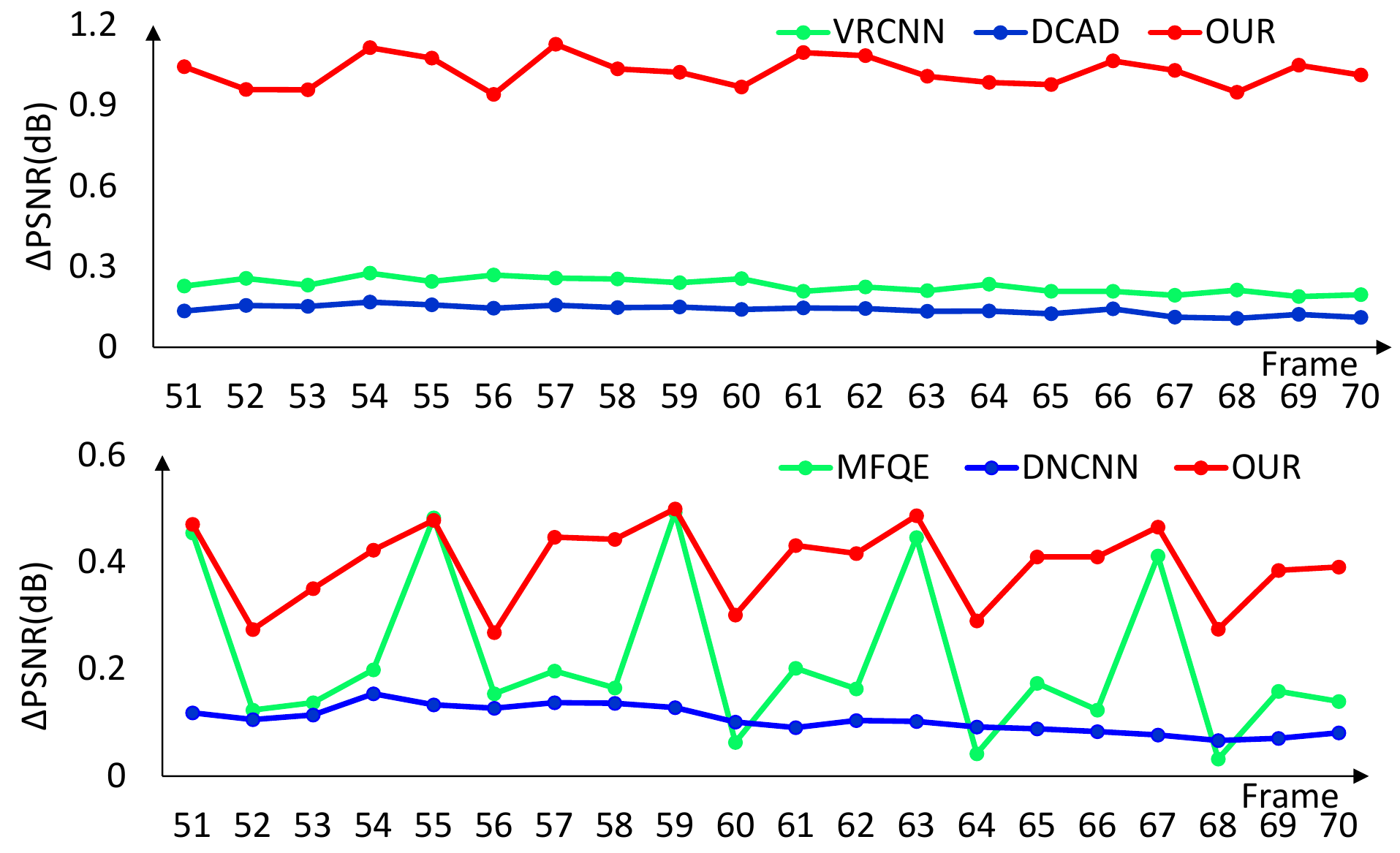}
\end{center}
\caption{\label{figure6} Comparison of $\Delta$PSNR (dB) curves for different methods under  AI (top) and LD (bottom) configurations.}
\end{figure}

In addition to the blocking and ringing artifacts of compressed video, quality fluctuations can also result in a degradation in the quality of experience \cite{QoE1,QoE}. In the experiment, we also compared the quality fluctuation of compressed video with comparison methods. As shown in Figure \ref{figure6}, we provide the $\Delta$PSNR results for 20 consecutive frames of the test video `BasketballDrill' under LD and AI configurations, respectively. From Figure  \ref{figure6}, we can see that the $\Delta$PSNR curve of our MGANet approach is always over the $\Delta$PSNR curves of comparison approaches. The PSNR fluctuation of our MGANet is obviously less than MFQE method. The curve of MFQE violently oscillates within the test frames, even lower than DnCNN approach for some frames that means our model is more robust than MFQE approach. To summarize, our MGANet approach is effective to mitigate the quality fluctuation of compressed video, meanwhile enhancing the compression video quality. %

\subsection{Robustness of MGANet Model for QPs}
\begin{figure}[htb]
  \centering
  \includegraphics[width=0.48\textwidth]{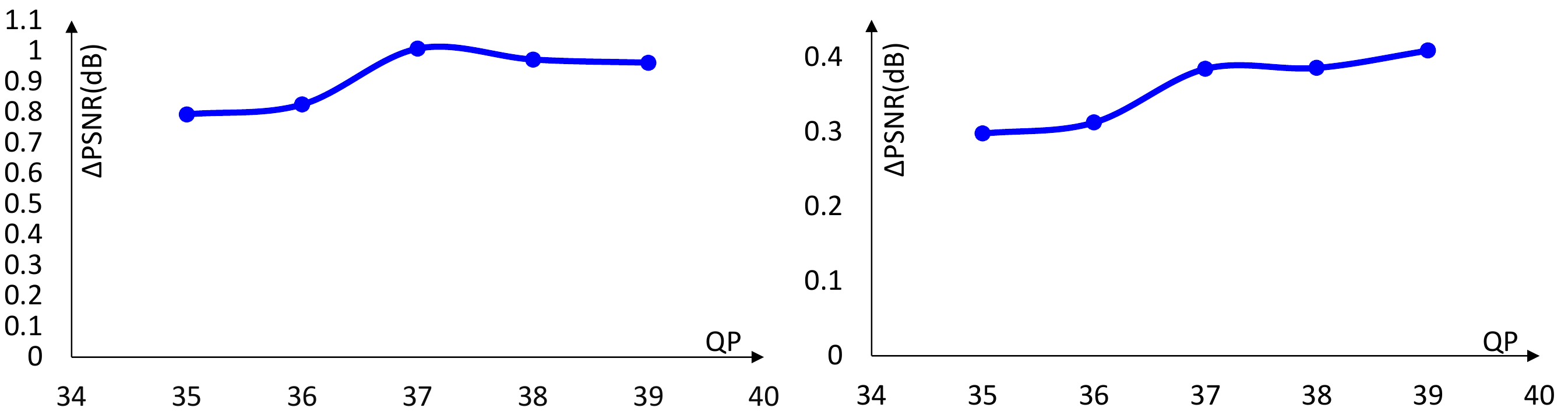}
  \caption{\label{Robustness}Average $\Delta$PSNR (dB) curves  of MGANet model for different QPs under AI (left) and LD (right) configurations} 
\end{figure}

In the above experiments, we have trained different models for different QPs. In practice, training a different model for each QP may be too costly. Therefore, we investigated the generalization capabilities of our model for different QPs. In this experiment, we use the models trained at QP 37, to test sequences at QPs 35, 36, 38 and 39, respectively.  As shown in Figure \ref{Robustness}, we can still observe a large $\Delta$PSNR (dB) can be reached by our MGANet approach, which shows the effectiveness of the training model for different QPs. Therefore, the number of training models required in practice may be much less than the number of possible QPs. Furthermore, since the higher QP corresponds to a lower bit-rate, the compression artifacts are usually more severe, the $\Delta$PSNR reduction of small QP is generally more than that of large QP.
\subsection{Ablation Study }
\begin{figure}[H]
  \centering
  \includegraphics[width=0.49\textwidth]{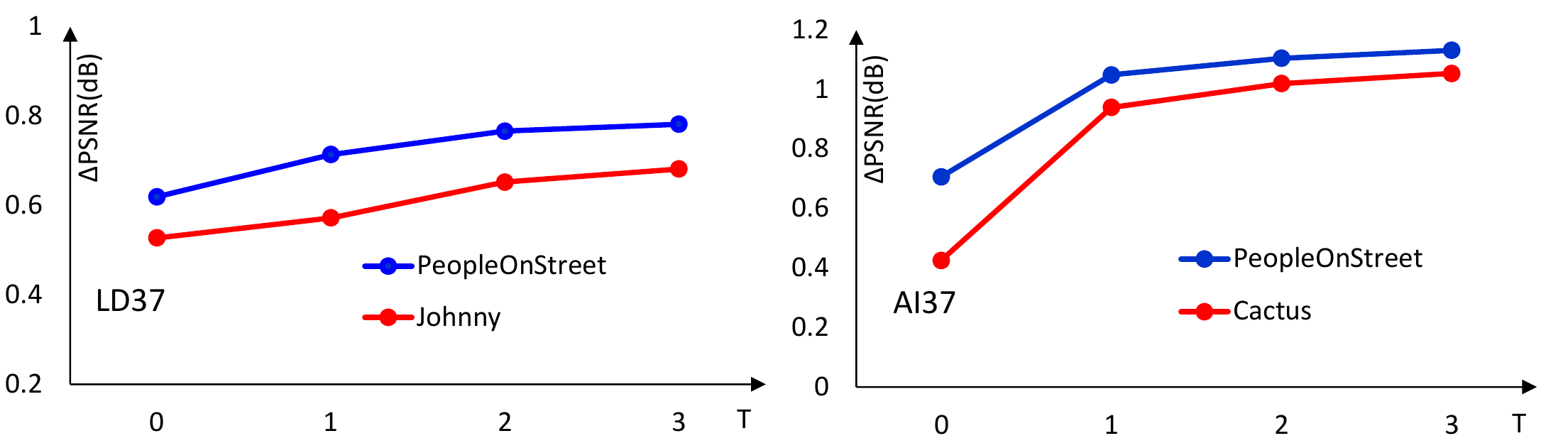}
  \caption{\label{radius}$\Delta$PSNR (dB) curves of MGANet model to test the influence of temporal neighborhood radius $T$} 
\end{figure}

\noindent{\bf{Temporal Neighborhood Radius}}  To investigate the influence of the temporal neighborhood radius $T$, we evaluate MGANet approach for different $T$, and the results are shown in Figure \ref{radius}. We only show the results of two sequences in the figure, and our experimental results show that other sequences have similar trends. As can be seen, performance grows as radius $T$ increases in general. But the performance gain seems to become marginal when $T\ge3$. This reflects the difficulty in exploring long-term temporal information, and is reserved for future study.  We regard $T = 1$ (that is, the nearest frame as a reference) as a balance between reconstruction quality and computational cost.

\begin{figure*}[htp]
\centering
\subfigure[ARCNN]
{
\begin{minipage}[b]{0.135\linewidth}
\includegraphics[width=1\linewidth]{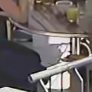} 
\includegraphics[width=1\linewidth]{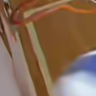}
\end{minipage}
}\hspace{-0.0095\linewidth}
\subfigure[VRCNN]
{
\begin{minipage}[b]{0.135\linewidth}
\includegraphics[width=1\linewidth]{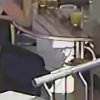}  
\includegraphics[width=1\linewidth]{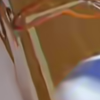}
\end{minipage}
} \hspace{-0.0135\linewidth}
\subfigure[DCAD]
{
\begin{minipage}[b]{0.135\linewidth}
\includegraphics[width=1\linewidth]{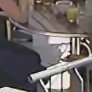} 
\includegraphics[width=1\linewidth]{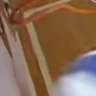}
\end{minipage}
}\hspace{-0.0095\linewidth}
\subfigure[DnCNN]
{
\begin{minipage}[b]{0.135\linewidth}
\includegraphics[width=1\linewidth]{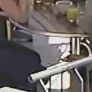} 
\includegraphics[width=1\linewidth]{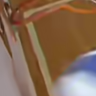}
\end{minipage}
}\hspace{-0.0095\linewidth}
\subfigure[MemNet]
{
\begin{minipage}[b]{0.135\linewidth}
\includegraphics[width=1\linewidth]{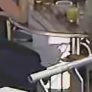} 
\includegraphics[width=1\linewidth]{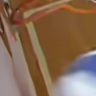}
\end{minipage}
}\hspace{-0.0095\linewidth}
\subfigure[Ours]
{
\begin{minipage}[b]{0.135\linewidth}
\includegraphics[width=1\linewidth]{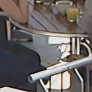} 
\includegraphics[width=1\linewidth]{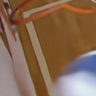}
\end{minipage}
}\hspace{-0.0095\linewidth}
\subfigure[Ground Truth]
{
\begin{minipage}[b]{0.135\linewidth}
\includegraphics[width=1\linewidth]{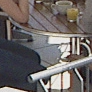} 
\includegraphics[width=1\linewidth]{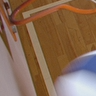}
\end{minipage}
}\hspace{-0.0095\linewidth}

\subfigure[ARCNN]
{
\begin{minipage}[b]{0.135\linewidth}
\includegraphics[width=1\linewidth]{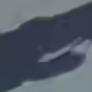} 
\includegraphics[width=1\linewidth]{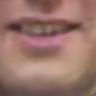}
\end{minipage}
}\hspace{-0.0095\linewidth}
\subfigure[DCAD]
{
\begin{minipage}[b]{0.135\linewidth}
\includegraphics[width=1\linewidth]{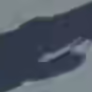} 
\includegraphics[width=1\linewidth]{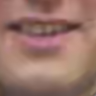}
\end{minipage}
}\hspace{-0.0095\linewidth}
\subfigure[DnCNN]
{
\begin{minipage}[b]{0.135\linewidth}
\includegraphics[width=1\linewidth]{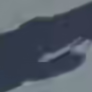} 
\includegraphics[width=1\linewidth]{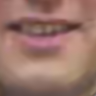}
\end{minipage}
}\hspace{-0.0095\linewidth}
\subfigure[MemNet]
{
\begin{minipage}[b]{0.135\linewidth}
\includegraphics[width=1\linewidth]{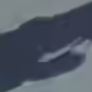} 
\includegraphics[width=1\linewidth]{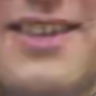}
\end{minipage}
}\hspace{-0.0095\linewidth}
\subfigure[MFQE]
{
\begin{minipage}[b]{0.135\linewidth}
\includegraphics[width=1\linewidth]{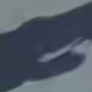} 
\includegraphics[width=1\linewidth]{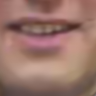}
\end{minipage}
}\hspace{-0.0095\linewidth}
\subfigure[Ours]
{
\begin{minipage}[b]{0.135\linewidth}
\includegraphics[width=1\linewidth]{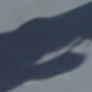} 
\includegraphics[width=1\linewidth]{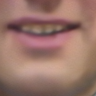}
\end{minipage}
}\hspace{-0.0095\linewidth}
\subfigure[Ground Truth]
{
\begin{minipage}[b]{0.135\linewidth}
\includegraphics[width=1\linewidth]{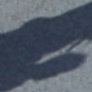} 
\includegraphics[width=1\linewidth]{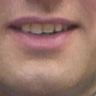}
\end{minipage}
}\hspace{-0.0095\linewidth}
\caption{\label{figurevisual}Subjective quality performance, images on the top two rows are from videos \textit{BQTerrace} and  \textit{BasketballDrill} at QP 37 under AI configuration, the bottom two rows are from videos \textit{PeopleOnStreet} and \textit{FourPeople} at QP 37 under LD configuration}
\end{figure*}

\begin{figure}[H]
  \centering
  \includegraphics[width=0.48\textwidth]{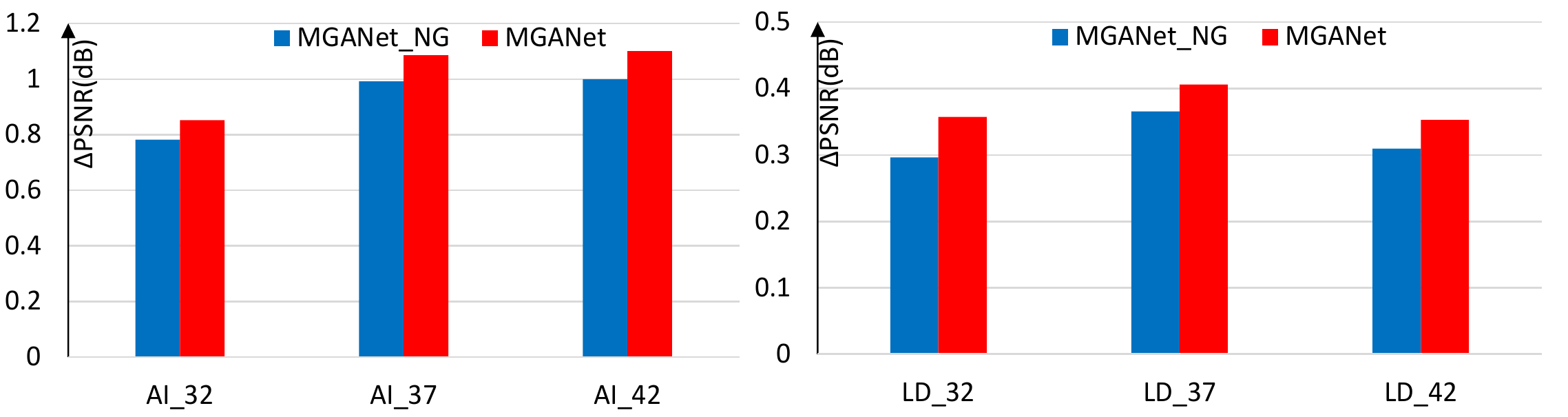}
  \caption{\label{mask}Average $\Delta$PSNR(dB) for the test set under LD and AI configurations to test the effects of guided map} 
\end{figure}

\noindent{\bf{Guided Map}}
Figure \ref{mask} presents the ablation study on the effects of Guided Map, compared to MGANet, MGANet$\_$NG  removes the guided map  component. We train 6  models at QPs 32, 37 and 42, under AI and LD configurations, respectively. The test results are shown in Figure \ref{mask}, we can see that all MGANet models are superior to MGANet$\_$NG models, which reveals the proposed intra-frame prior information plays an important role in the quality enhancement of compressed video. 

\noindent{\bf{BRCLSTM temporal encoder}}
To verify the influence of BRCLSTM, we study different treatments of the temporal dimension with early fusion, slow fusion and BCLSTM temporal encoder. The early fusion \cite{Early} collapses all temporal information in the first layer and the remaining operations are identical to those in a single frame network. While, slow fusion \cite{Slow} is to partially merge temporal information in a hierarchical structure. Compared to BRCLSTM, BCLSTM removes the residual connection.
We totally train 16 models with three fusion modes at QP 37 for $T=1$ (3-input frame, F3) and $T=2$ (5-input frame, F5) under AI and LD configuration, respectively. The results are shown in Table \ref{tab:BRCLSTM}, from the $\Delta$PSNR results we can see that our MGANet with BRCLSTM  outperforms the early fusion and slow fusion modes, which reveal the proposed BRCLST unit is useful for our MGANet approach. %
\renewcommand{\arraystretch}{2.2}
\begin{table}[!htp]
\centering
\setlength{\abovecaptionskip}{8pt}
\centering
\fontsize{9.0}{6.5}\selectfont
\caption{Ablation study on effects of BRCLSTM, average $\Delta$PSNR(dB) for the test set under LD and AI configurations.}
\label{tab:BRCLSTM}
\begin{tabular}{c|c|c|c|c}
\hline
Fusion Mode & AI$\_$F3 &AI$\_$F5 & LD$\_$F3 &LD$\_$F5 \\
\hline
Early Fusion &0.9134 & 0.8455 & 0.3687 & 0.3011 \\
\hline
Slow Fusion & 0.9470 & 0.8726 & 0.3672& 0.3036 \\
\hline
BCLSTM & 1.0035 & 1.1007 & 0.3393& 0.3563 \\
\hline
BRCLSTM & 1.0049 & 1.1123 & 0.4041 & 0.4106 \\
\hline
\end{tabular} 
\end{table}
\subsection{Running-time Evaluation}
Since our network is fully convolutional and LSTM-based temporal encoder, frames of arbitrary number and size can be fed in it as input, as long as GPU memory allows. Using our un-optimized PyTorch code, the F5 model takes about 18ms to process 5 input frames of size 416$\times$240  for one high quality frame output. Our method can be further accelerated to 15ms for F3  and 3ms for F1. That means our F3 model can generate about 67 high quality outputs per second with 3-input low quality frames. 
\subsection{Subjective Quality Performance}
Some subjective results are shown in Figure  \ref{figurevisual} for a more comprehensive and clearer comparison. For convenience, we intercept a portion of a frame and zoom in on them the same size. From Figure \ref{figurevisual}, we can see that the comparison methods are not efficient at removing some sharp blocking edges. Our proposed approach offers sharper edges, and some obvious blocking artifacts and ringings have been removed, it not only retains most of the structural information of the video content, but also restores some destroyed structures.  
\section{Conclusions}
In this paper, we systematically studied how to build an effective network for quality enhancement of compressed video, and proposed a guided attention network with multi-frame input.  Instead of explicitly calculating and compensating for motion between input frames, the BRCLSTM temporal encoder was designed  to implicitly discover inter-frame information. The guided encoder-decoder subnet was proposed to further enhance the quality of compressed video in spatial domain, and the guided map was used to guide our network to concentrate more on block boundary of compressed frame. Experimental results show that our MGANet significantly improves the quality of compressed video, far better than other state-of-the-art quality enhancement methods. The quantitative evaluation experiments and ablation studies demonstrated the robustness of our MGANet approach. This opens up new space for future exploration to use intra- and inter-frame prior information for quality enhancement of compressed video.

{\small
\bibliographystyle{ieee}
\bibliography{egbib}

\begin{thebibliography}{10}\itemsep=-1pt

\bibitem{Foi7}
V.~K. A.~Foi and K.~Egiazarian.
\newblock Pointwise shape-adaptive dct for high-quality denoising and
  deblocking of grayscale and color images.
\newblock {\em IEEE Transactions on Image Processing}, 16(5):1395 -- 1411,
  2007.

\bibitem{CTC}
F.~Bossen.
\newblock Common hm test conditions and software reference configurations.
\newblock {\em JCTVC-I1100, ITU-T SG16}, 2012.

\bibitem{Caballero}
J.~Caballero, C.~Ledig, A.~Aitken, A.~Acosta, J.~Totz, Z.~Wang, and W.~Shi.
\newblock Real-time video super-resolution with spatio-temporal networks and
  motion compensation.
\newblock In {\em CVPR}, 2017.

\bibitem{p1}
L.~Cavigelli, P.~Hager, and L.~Benini.
\newblock Cas-cnn: A deep convolutional neural network for image compression
  artifact suppression.
\newblock In {\em IJCNN}, 2015.

\bibitem{Chang8}
H.~Chang, M.~K. Ng, and T.~Zeng.
\newblock Reducing artifacts in jpeg decompression via a learned dictionary.
\newblock {\em IEEE Transactions on Image Processing}, 62(3):718 -- 728, 2014.

\bibitem{Cisco}
CVNI.
\newblock Cisco visual networking index: Global mobile data traffic forecast
  update, 2016-2021 white paper.
\newblock In {\em
  https://www.cisco.com/c/en/us/solutions/collateral/service-provider/visual-networking-index-vni/mobile-white-paper-c11-520862.html},
  2017.

\bibitem{Dai14}
Y.~Dai, D.~Liu, and F.~Wu.
\newblock A convolutional neural network approach for post-processing in hevc
  intra coding.
\newblock In {\em MMM}, pages 28 -- 39, 2017.

\bibitem{Dong9_15}
C.~Dong, Y.~Deng, C.~C. Loy, and X.~Tang.
\newblock Compression artifacts reduction by a deep convolutional network.
\newblock In {\em ICCV}, pages 576 -- 584, 2015.

\bibitem{ref5}
W.~Dong, G.~Shi, and X.~Li.
\newblock Nonlocal image restoration with bilateral variance estimation: A
  low-rank approach.
\newblock {\em IEEE Transactions on Image Processing}, 22(2):700 -- 711, 2013.

\bibitem{Foi}
A.~Foi, V.~Katkovnik, and K.~Egiazarian.
\newblock Pointwise shape-adaptive dct for high-quality denoising and
  deblocking of grayscale and color images.
\newblock {\em IEEE Transactions on Image Processing}, 16(5):1395--1411, 2007.

\bibitem{Guo10}
J.~Guo and H.~Chao.
\newblock Building dual-domain representations for compression artifacts
  reduction.
\newblock In {\em ECCV}, pages 628 -- 644, 2016.

\bibitem{yao1}
X.~He, Q.~Hu, X.~Han, X.~Zhang, C.~Zhang, and W.~Lin.
\newblock Enhancing hevc compressed videos with a partition-masked
  convolutional neural network.
\newblock In {\em ICIP}, pages 216--220, 2018.

\bibitem{QoE1}
Z.~He, Y.~K. Kim, and S.~K. Mitra.
\newblock Low-delay rate control for dct video coding via $\rho$-domain source
  modeling.
\newblock {\em IEEE Transactions on Circuits and Systems for Video Technology},
  11(8):928 -- 940, 2001.

\bibitem{Hochreiter22}
S.~Hochreiter and J.~Schmidhuber.
\newblock Long short-term memory.
\newblock {\em Neural Computation}, 9(8):1735 -- 1780, 1997.

\bibitem{QoE}
S.~Hu, H.~Wang, and S.~Kwong.
\newblock Adaptive quantizationparameter clip scheme for smooth quality in
  h.264/avc.
\newblock {\em IEEE Transactions on Image Processing}, 21(4):1911 -- 1919,
  2012.

\bibitem{Jancsary11}
J.~Jancsary, S.~Nowozin, and C.~Rother.
\newblock Compression artifacts reduction by a deep convolutional network.
\newblock In {\em ICCV}, pages 576 -- 584, 2015.

\bibitem{caffe}
Y.~Jia, E.~Shelhamer, J.~Donahue, S.~Karayev, J.~Long, R.~Girshick,
  S.~Guadarrama, and T.~Darrell.
\newblock Caffe: Convolutional architecture for fast feature embedding.
\newblock In {\em ACM Multimedia}, 2014.

\bibitem{kang}
L.~Kang, C.~C. Hsu, B.~Zhuang, C.~W. Lin, and C.~H. Yeh.
\newblock Learning-based joint super-resolution and deblocking for a highly
  compressed image.
\newblock {\em IEEE Transactions on Multimedia}, 17(7):921 -- 934, 2015.

\bibitem{Early}
A.~Kappeler, S.~Yoo, Q.~Dai, and A.~K. Katsaggelos.
\newblock Video super-resolution with convolutional neural networks.
\newblock {\em IEEE Transactions on Computational Imaging}, 2(2):109 -- 122,
  2016.

\bibitem{Slow}
A.~Karpathy, G.~Toderici, S.~Shetty, T.~Leung, R.~Sukthankar, and L.~Fei-Fei.
\newblock Large-scale video classification with convolutional neural networks.
\newblock In {\em CVPR}, 2014.

\bibitem{Adam}
Kingma and B.~Jimmy.
\newblock Adam: A method for stochastic optimization.
\newblock In {\em ICLR}, 2014.

\bibitem{Li1}
S.~Li, M.~Xu, Z.~Wang, and X.~Sun.
\newblock Optimal bit allocation for ctu level rate control in hevc.
\newblock {\em IEEE Transactions on Circuits and Systems for Video Technology},
  27(11):2409 -- 2424, 2017.

\bibitem{twostep1}
R.~Liao, X.~Tao, R.~Li, Z.~Ma, and J.~Jia.
\newblock Video superresolution via deep draft-ensemble learning.
\newblock In {\em ICCV}, pages 531--539, 2015.

\bibitem{List20}
P.~List, A.~Joch, J.~Lainema, G.~Bj$\phi$ntegaard, and M.~Karczewicz.
\newblock Adaptive deblocking filter.
\newblock {\em IEEE Transactions on Circuits and Systems for Video Technology},
  13(7):614 -- 619, 2003.

\bibitem{twostep2}
D.~Liu, Z.~Wang, Y.~Fan, X.~Liu, Z.~Wang, S.~Chang, and T.~Huang.
\newblock Robust video super-resolution with learned temporal dynamics.
\newblock In {\em ICCV}, 2017.

\bibitem{Liu21_14}
S.~Liu, N.~Yang, M.~Li, and M.~Zhou.
\newblock A recursive recurrent neural network for statistical machine
  translation.
\newblock In {\em ACL}, pages 1491 -- 1500, 2014.

\bibitem{Unet4}
Z.~Liu, R.~Yeh, X.~Tang, Y.~Liu, and A.~Agarwala.
\newblock Video frame synthesis using deep voxel flow.
\newblock In {\em ICCV}, 2017.

\bibitem{Unet1}
X.~Mao, C.~Shen, and Y.-B. Yang.
\newblock Image restoration using very deep convolutional encoder-decoder
  networks with symmetric skip connections.
\newblock In {\em NIPS}, 2016.

\bibitem{VVC}
J.~R. Ohm and G.~J. Sullivan.
\newblock Versatile video coding-towards the next generation of video
  compression.
\newblock In {\em PCS}, 2018.

\bibitem{dataset}
J.~R. Ohm, G.~J. Sullivan, H.~Schwarz, T.~K. Tan, and T.~Wiegand.
\newblock Comparison of the coding efficiency of video coding standards
  including high efficiency video coding (hevc).
\newblock {\em IEEE Transactions on Circuits and Systems for Video Technology},
  22(12):1669--1684, 2017.

\bibitem{Chuan2}
J.~S. Ren, Y.~Hu, Y.-W. Tai, C.~Wang, L.~Xu, W.~Sun, and Q.~Yan.
\newblock Look, listen and learn-a multimodal lstm for speaker identification.
\newblock In {\em Proceedings of the 30th AAAI Conference on Artificial
  Intelligence}, pages 3581--3587, 2016.

\bibitem{Unet2}
O.~Ronneberger, P.~Fischer, and T.~Brox.
\newblock U-net: Convolutional networks for biomedical image segmentation.
\newblock In {\em MICCAI}, pages 234 -- 241, 2015.

\bibitem{Shi23}
X.~Shi, Z.~Chen, H.~Wang, D.~Y. Yeung, W.~K. Wong, and W.~C. Woo.
\newblock Convolutional lstm network: A machine learning approach for
  precipitation nowcasting.
\newblock In {\em NIPS}, 2015.

\bibitem{Sullivan1}
G.~J. Sullivan, J.~R. Ohm, W.~J. Han, and T.~Wiegand.
\newblock Overview of the high efficiency video coding (hevc) standard.
\newblock {\em IEEE Transactions on Circuits and Systems for Video Technology},
  22(12):1649--1668, 2012.

\bibitem{Tai17}
Y.~Tai, J.~Yang, X.~Liu, and C.~Xu.
\newblock Memnet: A persistent memory network for image restoration.
\newblock In {\em ICCV}, 2017.

\bibitem{Unet5}
X.~Tao, H.~Gao, R.~Liao, J.~Wang, and J.~Jia.
\newblock Detail-revealing deep video super-resolution.
\newblock In {\em ICCV}, 2017.

\bibitem{Unet3}
X.~Tao, H.~Gao, Y.~Wang, X.~Shen, J.~Wang, and J.~Jia.
\newblock Scale-recurrent network for deep image deblurring.
\newblock In {\em CVPR}, 2018.

\bibitem{Chuan1}
C.~Wang, H.~Huang, X.~Han, and J.~Wang.
\newblock Video inpainting by jointly learning temporal structure and spatial
  details.
\newblock In {\em arXiv preprint arXiv:1806.08482}, 2018.

\bibitem{Wang15_17Chao}
T.~Wang, M.~Chen, and H.~Chao.
\newblock A novel deep learningbased method of improving coding efficiency from
  the decoder-end for hevc.
\newblock In {\em DCC}, 2017.

\bibitem{Wang12_16}
Z.~Wang, D.~Liu, S.~Chang, Q.~Ling, Y.~Yang, and T.~S. Huang.
\newblock D3: Deep dual-domain based fast restoration of jpeg-compressed
  images.
\newblock In {\em CVPR}, pages 2764 -- 2772, 2016.

\bibitem{Wiegand2}
T.~Wiegand, G.~J. Sullivan, G.~Bj$\phi$ntegaard, and A.~Luthra.
\newblock Overview of the h. 264/avc video coding standard.
\newblock {\em IEEE Transactions on Circuits and Systems for Video Technology},
  13(7):560 -- 576, 2003.

\bibitem{Yang16}
R.~Yang, M.~Xu, Z.~Wang, and T.~Li.
\newblock Multi-frame quality enhancement for compressed video.
\newblock In {\em CVPR}, pages 6664 -- 6673, 2018.

\bibitem{Yoo6}
S.~B. Yoo, K.~Choi, and J.~B. Ra.
\newblock Overview of the high efficiency video coding (hevc) standard.
\newblock {\em IEEE Transactions on Multimedia}, 16(6):1536--1548, 2014.

\bibitem{Zhai82}
G.~Zhai, W.~Zhang, X.~Yang, W.~Lin, , and Y.~Xu.
\newblock Efficient image deblocking based on postfiltering in shifted windows.
\newblock {\em IEEE Transactions on Circuits and Systems for Video Technology},
  18(1):122 -- 126, 2008.

\bibitem{Zhang18_16}
J.~Zhang, R.~Xiong, C.~Zhao, Y.~Zhang, S.~Ma, and W.~Gao.
\newblock Concolor: Constrained non-convex low-rank model for image deblocking.
\newblock {\em IEEE Transactions on Image Processing}, 25(3):1246 -- 1259,
  2016.

\bibitem{Zhang13_17}
K.~Zhang, W.~Zuo, Y.~Chen, D.~Meng, and L.~Zhang.
\newblock Beyond a gaussian denoiser: Residual learning of deep cnn for image
  denoising.
\newblock {\em IEEE Transactions on Image Processing}, 26(7):3142 -- 3155,
  2017.

\end{thebibliography}
}

\end{document}